\documentclass[11pt,a4paper]{article}

\usepackage{fontspec}

\usepackage{fontspec}
\usepackage{polyglossia}
\usepackage[table]{xcolor}
\usepackage{subcaption}
\usepackage{float}
\usepackage{array}
\usepackage{tabularx}
\usepackage[most]{tcolorbox}

\setdefaultlanguage{english}
\setotherlanguage{russian}
\usepackage{xcolor} 

\usepackage{csquotes}
\usepackage{natbib}
\usepackage[margin=1in]{geometry}
\usepackage{microtype}
\usepackage{setspace}
\usepackage{amsmath, amssymb}
\usepackage{booktabs}
\usepackage{siunitx}

\usepackage{graphicx}
\usepackage{caption}
\usepackage{subcaption}
\usepackage{float} 
\usepackage{tikz}
\usetikzlibrary{trees, positioning, arrows.meta, shapes.geometric, shadows, shapes.misc, calc}
\usepackage{graphicx}
\usepackage{xcolor}

\usepackage[hidelinks]{hyperref}

\title{The Devil is in the Details -- From OCR for Old Church Slavonic to Purely Visual Stemma Reconstruction}
\author{Armin Hoenen \\
Department of Empirical Linguistics \\
University of Frankfurt}

\date{\today}

\begin{document}

\maketitle

\abstract{\textit{The age of artificial intelligence has brought many new possibilities and pitfalls in many fields and tasks. The devil is in the details, and those come to the fore when building new pipelines and executing small practical experiments. OCR and stemmatology are no exception. The current investigation starts comparing a range of OCR-systems, from classical over machine learning to LLMs, for roughly 6,000 characters of late handwritten church slavonic manuscripts from the 18th century. Focussing on basic letter correctness, more than 10 CS OCR-systems among which 2 LLMs (GPT5 and Gemini3-flash) are being compared. Then, post-processing via LLMs is assessed and finally, different agentic OCR architectures (specialized post-processing agents, an agentic pipeline and RAG) are tested. With new technology elaborated, experiments suggest, church slavonic CER for basic letters may reach as low as 2-3\% but elaborated diacritics could still present a problem. How well OCR can prime stemmatology as a downstream task is the entry point to the second part of the article which introduces a new stemmatic method based solely on image processing. Here, a pipeline of automated visual glyph extraction, clustering and pairwise statistical comparison leading to a distance matrix and ultimately a stemma, is being presented and applied to two small corpora, one for the church slavonic Gospel of Mark from the 14th to 16th centuries, one for the Roman de la Rose in French from the 14th and 15th centuries. Basic functioning of the method can be demonstrated.\footnote{This is a draft-version of the article.}}}
\tableofcontents

\section{Prehistory: A trinity of experiments in Venice}
In \citet{hoenen:2025}, visual models (YOLO, self-trained \& GPT) are being used to graphically identify early stemmata in digital collections, a kind of distant reading. A second experiment aimed to extract witness relations from textual sources (ideally early prints) from before the widespread use of stemmata. The goal would be a project clarifying the prehistory of stemmatology.

\noindent Whilst LLMs were used for the latter experiment, they were tested and optimized with a synthetic modern English digital dataset with good success. The applicability to historically relevant languages such as Latin, was then tested briefly on one text.

Here, faulty OCR was given to GPT in order to do text extraction. OCR errors rarely interfere with relation extraction unless they coincide with structurally relevant elements such as manuscript identifiers. In practice, the more probable failure mode is not misclassification but retrieval failure, where relevant relations are simply not detected. For more in depth discussions and a general perspective on the impact of OCR-errors in Information Retrieval, consider \cite{bazzo:2020}.

Recent literature supports the potential of LLM-based OCR post-correction. \cite{greif:2025}, for instance, demonstrate that even heavily degraded historical print (Fraktur, Antiqua, 18th cent.) can be normalized to near-editorial quality, achieving character error rates between one and two percent by using an OCR model plus post-correction via an LLM without task-specific fine-tuning. This naturally raises the question of whether similar success can be achieved for more difficult material such as handwritten Old Church Slavonic.

If this works well for historical print in Latin with a CER of $<1\%$, what is the applicability for (Old) Church Slavonic handwriting. Where are existing models and will LLM-post-correction help for this scenario?

\section{OCR for Slavonic}
We make several experiments with extraction engines, the aim of which is under more to determine the best or at least feasible solutions for handwritten church slavonic text.

\subsection{The Valamo Test Corpus}
A small corpus was constructed from three handwritten manuscripts originating from the Valamo monastery in the eighteenth century (all that was available February/March 2026).

\begin{itemize}
\itemsep0em
    \item XII.254 -- \textit{Akathistos to John the Baptist} \\
          Valamo Monastery, 1760s
    \item XII.179 -- \textit{Symeon the New Theologian} \\
          Slavic translation copied in 1786
    \item XII.172 -- \textit{Prayer and notebook collection} \\
          Monastic manuscript, 1787
\end{itemize}

\noindent For each manuscript, the first three pages were screenshot and saved as png. The transcriptions from the page were saved accordingly. In all this equalled $6000$ characters.

\subsection{Models tested}

\begin{table}[H]
\centering
\small
\begin{tabularx}{\textwidth}{|p{2.5cm}|p{2.2cm}|X|p{1.8cm}|p{2.5cm}|}
\hline
\textbf{Model} & \textbf{Type} & \textbf{Data / Target} & \textbf{Perf} & \textbf{Local / Cloud (Req.)} \\
\hline

Gemini 3 Flash & VLM & Multimodal web-scale & $\sim$0.115 & Cloud only (API, no local) \\
\hline

GPT-5 & LLM & Multilingual + vision & $\sim$0.147 & Cloud only (API) \\
\hline

Transkribus CS HTR 3 & HTR & OCS manuscripts (2.6k pp.) & CER $\sim$3.3\% & Cloud (Transkribus; no local) \\
\hline

Transkribus (Ogorodok, Ukrainian CS \& PMP, Ausbul, VKS2, Ostromir, Troitskiy) & HTR & Various OCS / Cyrillic MSS & $\sim$3–4\% & Cloud (Transkribus) \\
\hline

Qwen3-VL-8B OCS & VLM (8B) & Fine-tuned OCS & n/a & Local (GPU $\geq$16GB VRAM) \\
\hline

LightOnOCR-2-1B OCS & VLM (1B) & OCS line images & CER $\sim$5.0\% & Local (GPU $\geq$8–12GB) \\
\hline

Kraken Cyr002 & HTR & Historical Cyrillic & n/a & Local (CPU/GPU, low req.) \\
\hline

TrOCR (kazars24) & Transformer OCR & Cyrillic handwriting & CER $\sim$25\% & Local (GPU recommended $\geq$8GB) \\
\hline

Tesseract (rus, ell, eng) & OCR (LSTM) & Modern Cyrillic/Greek/Latin & poor (OCS) & Local (CPU, very low req.) \\
\hline

\end{tabularx}
\caption{Overview of OCR/HTR models for Old Church Slavonic (OCS) tested either punctually or systematically.}
\end{table}

\normalsize
Upon inspection, TrOCR produced no useful output, as it is more suitable to modern handwriting. Kraken and Transkribus Ausbul, Combined Full VKS 2, Ostromir and Troitskiy on a few test images performed a lot worse than the other models. These models have been excluded.

\subsubsection{Tesseract Baseline}
As a possible baseline and as such an anchor for comparison with older literature and other approaches, tesseract was tried. If a post-correction by LLM could restore texts, some extraction tasks or even corrections of standard pdfs and so on would be feasible. Unfortunately, the tests witnesses such a high CER, that this is implausible and an extraction via the LLM directly from the image more useful. 

\begin{table}[H]
\centering
\begin{tabular}[H]{lccccccc}
\toprule
Page & Ru & El & En & Ru+El & Ru+En & En+el & Ru+El+En \\
\midrule
\rowcolor{gray!15}
1 & \textbf{0.6026} & 0.9007 & 0.9227 & 0.6777 & 0.8079 & 0.8918 & 0.8212 \\
\rowcolor{gray!15}
2 & \textbf{0.6304} & 0.9149 & 0.9420 & 0.7518 & 0.8315 & 0.9330 & 0.8442 \\
\rowcolor{gray!15}
3 & \textbf{0.5886} & 0.8927 & 0.8998 & 0.6315 & 0.7245 & 0.8962 & 0.7352 \\
4 & \textbf{0.6312} & 0.8915 & 0.8980 & 0.6941 & 0.8308 & 0.8850 & 0.8178 \\
5 & \textbf{0.7571} & 0.9199 & 0.9121 & 0.8398 & 0.8605 & 0.9070 & 0.8630 \\
6 & \textbf{0.7282} & 0.8859 & 0.8900 & 0.7988 & 0.8382 & 0.8880 & 0.8382 \\
\rowcolor{gray!15}
7 & \textbf{0.6281} & 0.9105 & 0.9456 & 0.6930 & 0.8140 & 0.9193 & 0.8018 \\
\rowcolor{gray!15}
8 & \textbf{0.8419} & 1.2341 & 1.3116 & 0.9535 & 1.1070 & 1.2589 & 1.1411 \\
\rowcolor{gray!15}
9 & \textbf{0.5262} & 0.8950 & 0.9112 & 0.5962 & 0.7147 & 0.8896 & 0.7322 \\
\midrule
Mean & \textbf{0.6594} & 0.9384 & 0.9592 & 0.7374 & 0.8366 & 0.9410 & 0.8439 \\
\bottomrule
\end{tabular}
\caption{Results CER of tesseract on the images from the Valamo manuscripts.}
\end{table}

\noindent As one can see, the error rates are unfeasibly high. A restoration of the original text is not feasible even though some letters may have been correctly read. The combination with Greek and English (Latin alphabet) was attempted in order for some Old Cyrillic letters to be in the set of possible candidate letters and indeed small Omega from Greek and i from Latin were sometimes matched (Ukrainian and other tesseract languages were not used).

\subsection{Metric(s) used}
As metrics, Levenshtein distance is used as well as CER (Character error rate). Additionally, before computing these values in the default setting, texts were lowercased, all spaces and line breaks removed and diacritical marks, too. Results for and with them are marked.

\subsection{Visual Preprocessing}

Some visual processings via python scripts, commandline tools and the tool upscayl were attempted. 
\begin{itemize}
\itemsep0em
    \item via an upscaling algorithm the images were enhanced (4x ultramix balanced).
    \item conversion into grayscale was done and subsequent removal of dots: the resulting images did not improve OCR
    \item red ink letters were slightly redrawn with black lines: this could help the models which were struggling to extract portions of red letters, especially with few contrast to the background. However, the tested image deteriorated slightly and the script was not used for the experiments, since it would have supported some models more than others. However, some lines were recognized slightly better.
\end{itemize}

\noindent A first attempt enhancing contrast on red letters on darker brown ground helped Transkribus a bit.

\begin{figure}[H]
\centering
\begin{center}
    \includegraphics[scale=0.3]{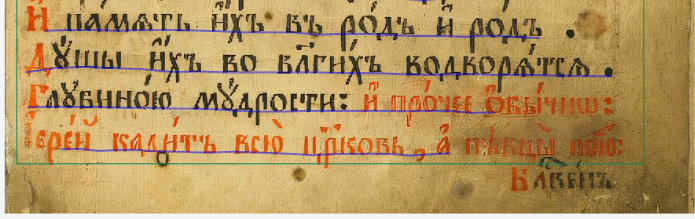}
        \includegraphics[scale=0.3]{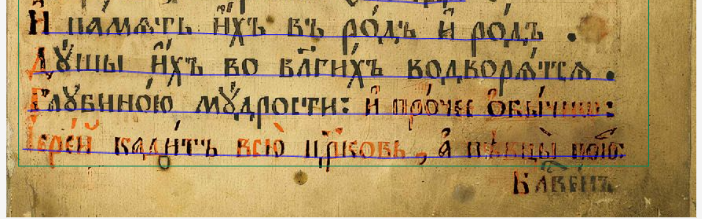}
\end{center}
\caption{Here, Red2Black made the last word appear in the transcript but deteriorated the second last.}
\end{figure}

The results of upscaling were, which is congruent with the literature, very positive. In the following table, the models suffixed with a 0 were the models on the non-upscaled images.
\begin{table}[H]
\centering
\begin{tabular}{lcccccccc}
\toprule
Page & T gcsh & T 0 & T r2b & T bw & Qwen3 & Q 0 & LightOn & L 0 \\
\midrule
\rowcolor{gray!15}
1 & 0.130 & 0.159 & -- & 0.349 & 0.088 & 0.214 & \textbf{0.055} & 0.126 \\
\rowcolor{gray!15}
2 & 0.096 & 0.159 & -- & -- & 0.072 & 0.147 & \textbf{0.031} & 0.069 \\
\rowcolor{gray!15}
3 & 0.172 & 0.172 & 0.179 & -- & 0.070 & 0.286 & \textbf{0.061} & 0.168 \\
4 & 0.349 & 0.289 & -- & -- & 0.134 & 0.308 & \textbf{0.095} & 0.176 \\
5 & 0.357 & 0.245 & -- & -- & 0.152 & 0.305 & \textbf{0.065} & 0.124 \\
6 & 0.510 & 0.199 & -- & -- & 0.191 & 0.402 & \textbf{0.112} & 0.154 \\
\rowcolor{gray!15}
7 & 0.230 & 0.291 & -- & -- & 0.177 & 0.325 & \textbf{0.075} & 0.191 \\
\rowcolor{gray!15}
8 & 0.093 & 0.212 & -- & -- & 0.110 & 0.191 & \textbf{0.051} & 0.081 \\
\rowcolor{gray!15}
9 & 0.092 & 0.245 & -- & -- & 0.149 & 0.308 & \textbf{0.077} & 0.118 \\
\midrule
Sum & 0.209 & 0.219 & 0.179 & 0.349 & 0.127 & 0.274 & \textbf{0.068} & 0.132 \\
\bottomrule
\end{tabular}
\caption{Base OCR results on non-upscaled (suffix 0) and upscaled images for base letters.}
\end{table}
\noindent Upscayl has a significantly positive effect, but, the Transkribus model deteriorated for images 4, 5 and 6 from the second manuscript (but upscaling helped them on the others). This is probably due to some enhancing effect of the border as artifact even though even with the non-upscaled images, the model partly transcribed the border. In order to be able to compare all models, the rest of the experiments was conducted on the upscaled images if not otherwise stated.

\begin{table}[H]
\centering
\begin{tabular}{lcccccccc}
\toprule
Page & T gcsh & T 0 & T r2b & T bw & Qwen3 & Q 0 & LightOn & L 0 \\
\midrule
\rowcolor{gray!15}
1 & 0.895 & 0.884 & -- & 0.895 & 0.863 & 0.800 & \textbf{0.737} & 0.832 \\
\rowcolor{gray!15}
2 & 0.908 & 0.900 & -- & -- & 0.775 & 0.875 & \textbf{0.742} & 0.750 \\
\rowcolor{gray!15}
3 & 0.880 & 0.887 & 0.902 & -- & 0.782 & 0.827 & \textbf{0.774} & 0.812 \\
4 & 0.846 & 0.856 & -- & -- & 0.740 & 0.750 & \textbf{0.721} & 0.731 \\
5 & 0.944 & 0.911 & -- & -- & 0.889 & 0.889 & \textbf{0.844} & 0.867 \\
6 & 0.943 & 0.868 & -- & -- & 0.830 & 0.877 & \textbf{0.811} & \textbf{0.811} \\
\rowcolor{gray!15}
7 & 0.790 & 0.919 & -- & -- & 0.758 & 0.831 & \textbf{0.718} & 0.815 \\
\rowcolor{gray!15}
8 & 0.849 & 0.905 & -- & -- & 0.841 & 0.857 & \textbf{0.817} & 0.841 \\
\rowcolor{gray!15}
9 & \textbf{0.889} & 0.903 & -- & -- & 0.903 & 0.896 & \textbf{0.889} & 0.896 \\
\midrule
Sum & 0.880 & 0.893 & 0.902 & 0.895 & 0.820 & 0.846 & \textbf{0.786} & 0.819 \\
\bottomrule
\end{tabular}
\caption{Base OCR results on non-upscaled (suffix 0) and upscaled images solely for diacritical marks.}
\end{table}
\noindent While upscaling improved results for diacritics (breathing marks, titlo, stress marks etc.) as by unicodedata.category 'Mn' in the LightOn model, for Qwen3 and Transkribus, the upscaling was sometimes hurting transcription.

\subsection{RESULTS: base OCR}
For obtaining a base OCR, several systems, as elicited, were tested. The three basic model families are large VLMs/LLMs, OCR-task models and VLMs fine-tuned for some form of church slavonic cyrillic. The results may be interpreted also as baselines for any subsequent post-correction and so forth. Performance and cost are not monitored. The main focus here is on accuracy, not economic aspects, which have been researched in other papers. The Qwen was run on CPU and quantized to 16Bit due to private hardware limitations, this model took longest for the generation of the results. 
\begin{table}[H]
\centering
\begin{tabular}{lccccccc}
\toprule
Page & Gem & GPT & T ukr & T gcsh & T ogo & Qwen3 & LightOn \\
\midrule
\rowcolor{gray!15}
1 & 0.079 & 0.161 & 0.071 & 0.130 & 0.126 & 0.088 & \textbf{0.055} \\
\rowcolor{gray!15}
2 & 0.053 & 0.188 & 0.058 & 0.096 & 0.085 & 0.072 & \textbf{0.031} \\
\rowcolor{gray!15}
3 & 0.079 & 0.261 & 0.116 & 0.172 & 0.154 & 0.070 & \textbf{0.061} \\
4 & 0.117 & 0.388 & 0.347 & 0.349 & 0.388 & 0.134 & \textbf{0.095} \\
5 & 0.088 & 0.238 & 0.326 & 0.357 & 0.390 & 0.152 & \textbf{0.065} \\
6 & 0.154 & 0.334 & 0.504 & 0.510 & 0.539 & 0.191 & \textbf{0.112} \\
\rowcolor{gray!15}
7 & 0.104 & 0.277 & 0.193 & 0.230 & 0.214 & 0.177 & \textbf{0.075} \\
\rowcolor{gray!15}
8 & 0.132 & 0.225 & \textbf{0.036} & 0.093 & 0.073 & 0.110 & 0.051 \\
\rowcolor{gray!15}
9 & 0.101 & 0.187 & \textbf{0.059} & 0.092 & 0.093 & 0.149 & 0.077 \\
\midrule
Sum & 0.101 & 0.247 & 0.172 & 0.209 & 0.210 & 0.127 & \textbf{0.068} \\
\bottomrule
\end{tabular}
\caption{Basic OCR results for different models on upscaled images and for normalized text.}
\end{table}
Overall best results were for LightOn and Gemini. All Transkribus models suffered from the fact that they read artistic borders of the second manuscript into nonsensical text fragments. 

Results without any normalization:
\begin{table}[H]
\centering
\begin{tabular}{lccccccc}
\toprule
Page & Gem & GPT & T ukr & T gcsh & T ogo & Qwen3 & LightOn \\
\midrule
\rowcolor{gray!15}
1 & 0.222 & 0.306 & \textbf{0.141} & 0.276 & 0.211 & 0.241 & 0.232 \\
\rowcolor{gray!15}
2 & 0.179 & 0.316 & \textbf{0.084} & 0.251 & 0.157 & 0.228 & 0.166 \\
\rowcolor{gray!15}
3 & 0.250 & 0.382 & \textbf{0.173} & 0.305 & 0.219 & 0.234 & 0.232 \\
4 & \textbf{0.234} & 0.474 & 0.406 & 0.441 & 0.456 & 0.295 & 0.259 \\
5 & \textbf{0.228} & 0.354 & 0.406 & 0.479 & 0.490 & 0.324 & 0.241 \\
6 & 0.302 & 0.431 & 0.538 & 0.588 & 0.583 & 0.319 & \textbf{0.270} \\
\rowcolor{gray!15}
7 & 0.258 & 0.373 & 0.261 & 0.326 & 0.301 & 0.284 & \textbf{0.199} \\
\rowcolor{gray!15}
8 & 0.255 & 0.323 & \textbf{0.087} & 0.221 & 0.159 & 0.237 & 0.187 \\
\rowcolor{gray!15}
9 & 0.231 & 0.290 & \textbf{0.117} & 0.219 & 0.173 & 0.267 & 0.212 \\
\midrule
Sum & 0.240 & 0.356 & 0.228 & 0.330 & 0.287 & 0.267 & \textbf{0.219} \\
\bottomrule
\end{tabular}
\caption{Basic OCR results raw.}
\end{table}
Results only for diacritical marks:

\begin{table}[H]
\centering
\begin{tabular}{lccccccc}
\toprule
Page & Gem & GPT & T ukr & T gcsh & T ogo & Qwen3 & LightOn \\
\midrule
\rowcolor{gray!15}
1 & 0.705 & 0.916 & \textbf{0.242} & 0.895 & 0.411 & 0.863 & 0.737 \\
\rowcolor{gray!15}
2 & 0.783 & 0.983 & \textbf{0.142} & 0.908 & 0.333 & 0.775 & 0.742 \\
\rowcolor{gray!15}
3 & 0.910 & 0.970 & \textbf{0.263} & 0.880 & 0.331 & 0.782 & 0.774 \\
4 & 0.827 & 0.952 & \textbf{0.471} & 0.846 & 0.519 & 0.740 & 0.721 \\
5 & 0.856 & 0.867 & \textbf{0.456} & 0.944 & 0.489 & 0.889 & 0.844 \\
6 & 0.981 & 1.000 & \textbf{0.604} & 0.943 & 0.642 & 0.830 & 0.811 \\
\rowcolor{gray!15}
7 & 0.976 & 0.984 & \textbf{0.274} & 0.790 & 0.379 & 0.758 & 0.718 \\
\rowcolor{gray!15}
8 & 0.984 & 0.984 & \textbf{0.198} & 0.849 & 0.341 & 0.841 & 0.817 \\
\rowcolor{gray!15}
9 & 0.861 & 0.917 & \textbf{0.250} & 0.889 & 0.361 & 0.903 & 0.889 \\
\midrule
Sum & 0.881 & 0.955 & \textbf{0.311} & 0.880 & 0.414 & 0.820 & 0.786 \\
\bottomrule
\end{tabular}
\caption{Basic OCR results for diacritical marks.}
\end{table}
\noindent Two Transkribus models were the only ones, dealing well with the complex diacritical marks of those manuscripts, even though even they struggled. Diacritical marks are on the one hand smaller, artistically probably more diverse and also presumably less standardized in their usage than the base letters, which makes them extremely difficult for OCR systems. 

\subsection{Substitutions, Deletions and Insertions}
\begin{table}[H]
\centering
\includegraphics[scale=0.5]{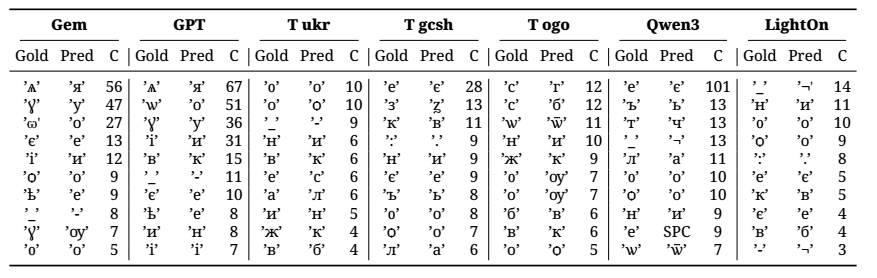}
\caption{Top OCR confusions.}
\end{table}

\noindent Whilst most substitutions are due to visual confusion, the two large LLMs show very strong modernization (editorial). Ancient letters are constantly replaced by modern ones maybe with a tendency towards those used in modern Russian as this must be the most frequent cyrillic language in their training data. 
Thus, if the extraction prompt was optimized suppressing such urge, what would happen?
Their previous base prompt: ``Extract the text from this image.''\footnote{In the prompting literature, there is a robust trend towards using short and concise prompts, under more to reduce the so-called cognitive load, that is the complexity of the task.} was extended to: ``Extract the text from this image, do not confuse ѧ \& я; ꙋ \& у and ѡ \& о, retain the old characters. Do not modernize or insert/delete.'' This featured both exemplary mentions of the characters not to be confused and meta-linguistic terminology.

\begin{table}[H]
\centering
\begin{tabular}{lcccccc}
\toprule
Page & Gem & Gem 2 & GPT & GPT 2 & T ukr & LightOn \\
\midrule
\rowcolor{gray!15}
1 & 0.079 & \textbf{0.055} & 0.161 & 0.097 & 0.071 & \textbf{0.055} \\
\rowcolor{gray!15}
2 & 0.053 & 0.036 & 0.188 & 0.168 & 0.058 & \textbf{0.031} \\
\rowcolor{gray!15}
3 & 0.079 & \textbf{0.020} & 0.261 & 0.156 & 0.116 & 0.061 \\
4 & 0.117 & 0.154 & 0.388 & 0.223 & 0.347 & \textbf{0.095} \\
5 & 0.088 & \textbf{0.052} & 0.238 & 0.137 & 0.326 & 0.065 \\
6 & 0.154 & \textbf{0.060} & 0.334 & 0.290 & 0.504 & 0.112 \\
\rowcolor{gray!15}
7 & 0.104 & \textbf{0.053} & 0.277 & 0.258 & 0.193 & 0.075 \\
\rowcolor{gray!15}
8 & 0.132 & 0.040 & 0.225 & 0.181 & \textbf{0.036} & 0.051 \\
\rowcolor{gray!15}
9 & 0.101 & 0.083 & 0.187 & 0.159 & \textbf{0.059} & 0.077 \\
\midrule
Sum & 0.101 & \textbf{0.061} & 0.247 & 0.186 & 0.172 & 0.068 \\
\bottomrule
\end{tabular}
\caption{Basic OCR results with optimized prompt for LLMs.}
\end{table}

\noindent This small optimization made Gemini 2 the strongest model. GPT also strongly improved. Now, if the Gemini model is the base model and has an optimized prompt for extraction, the question arises whether post-correction at least self-post-correction is still worthwhile. But before going to this, a look at model similarities and a small side-experiment. 

\subsection{Model similarities}
Based on the vectors of substitutions, insertions and deletions, a model similarity tree was computed using UPGMA. 
\begin{figure}[H]
\centering
\begin{minipage}{0.8\textwidth}
    \centering
    \includegraphics[width=\linewidth]{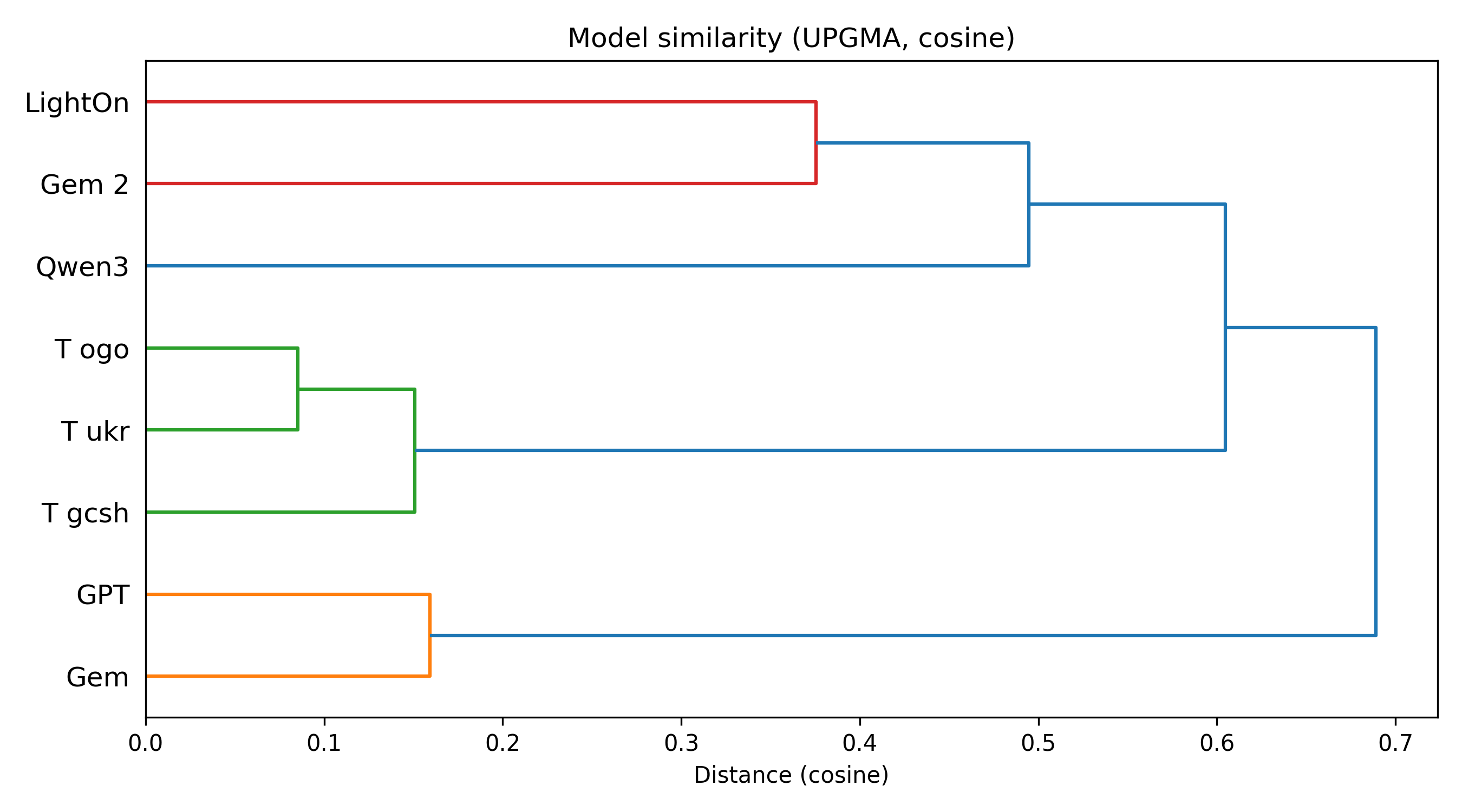}
\end{minipage}
\caption{Model similarities. Base Models ($\rightarrow$ families cluster such as LLMs, Transkribus models, fine-tuned models), the only exception being Gem 2.}
\end{figure}
\noindent Model families clearly cluster, but the prompt optimization pushes Gem 2 into a very different direction showing how much influence prompt optimization can have. 

\subsection{Single letters vs. letters in context}
In order to see, for instance, how strong the editorial urge of larger LLMs is dependent on context, whilst the basic visual extraction works fine, the images were automatically cut into glyphs and inconsistent images deleted, the other ones mapped to their gold letter.
Then, GPT4o\footnote{The author has access to a Plus-Account and the API for GPT, and a basic plan for Gemini, which contrained model choice at several points in this paper. Some results were obtained using the web interface, some through the API. Sometimes, the LLM answered in Russian or even Church Slavonic, a well-known case of language confusion. Therefore when using the API, a byphrase 'Provide only the OCR-result.' or an instruction to use JSON as machine-readable answer format are useful.} was used to ocr the single letter images and the statistics on confusions of single letters and letters-in-context from the previous experiment (GPT5) were compared. 

LightOn, when given the single glyphs, was not able to read single glyphs, it gave more than one letter and partly wild hallucinations, for a single ꙗ it gave `и саⷣ ⷪ' for instance. This may however be a method to trigger basic hallucination patterns for the model.  

\begin{table}[H]
\centering
\includegraphics[scale=0.5]{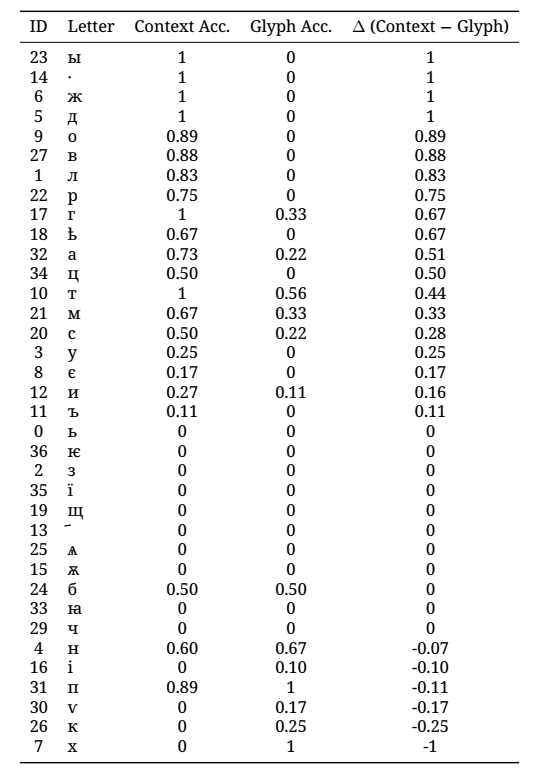}
\caption{Per-letter comparison for many letters, context-based vs. glyph-based OCR accuracy.}
\end{table}

\noindent Context-based recognition (GPT5) outperforms glyph-only recognition (GPT4-o) for many characters, especially ambiguous or variably written ones. This indicates that linguistic context is critical for resolving paleographic variation and noisy forms. However, a small subset of characters (e.g., н, п, х) performs better under glyph-based recognition, suggesting that for visually distinctive or abbreviation-prone forms, direct glyph modeling can be more reliable. н in context was 3 times matched, once misread as ѣ and as м once, as a glyph it was twice matched and once misread as ѳ which also features the horizontal line in the middle and which when cut in a bad way can be visually quite close. Glyph confusions are indeed mainly visual and therefore more limited in confusion possibilities, more predictable. Contextual confusions are more diverse and probably easier to post-correct when the visual difference is stark.  

\subsection{LLM Post-correction - modes}

The first thing to check was the configuration or mode, the LLM would postcorrect in. Here, the Qwen model results were used in an early stage of the experiment to reveal that a prompt in thinking-mode performed worse than the post-correction prompt alongside the uploaded image. Post correction without the image was worse (although here punctually in Gemini still better than the thinking mode result). 

\begin{table}[H]
\centering
\small
\begin{tabular}{l c}
\toprule
Pipeline (QWEN) & Mean (LEV) \\
\midrule
Gem & 28.33 \\
GPT & 46.67 \\
GemThink & 31.33 \\
GPTThink & 40.33 \\
Gem+Img & \textbf{22.33} [1.] \\
GPT+Img & 29 \\
\bottomrule
\end{tabular}
\caption{Modes of GPT and Gemini available in early 2026.}
\end{table}
\normalsize

\subsubsection{Post correction and web searches}
Another mode tried was the `deep research' mode, where the LLM was given the Image, the OCR and this prompt:\\
\begin{tcolorbox}[
    colback=gray!5, 
    colframe=gray!50, 
    arc=2pt, 
    boxrule=0.5pt, 
    left=10pt, 
    right=10pt, 
    top=8pt, 
    bottom=8pt
]
    \small \textbf{Prompt:} \\
Given this image and this OCR:

<transcript>

analyse the OCR and correct it using web search for disambiguation in case some feature is not entirely clear.''
\end{tcolorbox}

An obvious advantage of this mode is, that it can find previous published OCRs of the same text/image. In this case especially, the LLM could in theory find the Valamo website. However, such a mode can be explicitly instructed to use certain web ressources. Gemini did not identify the Valamo page, but used the page \url{azbuka.ru}.
The deep research modes are specialized agentic pipelines, built for web research. They generate a report and give their sources. These modes are costly and have been used here only in limited scenarios. GPT additioally offers a mode "web search" which is a mode making the model search on the web.

For GPT, the second image was uploaded together with the Qwen OCR. The web serarch mode led to a significant deterioration of CER (qwen basic: 0.134, web search 0.284) whilst the deep research more improved the CER by a large margin (deep research: 0.093). 
For Gemini, the mode deep research was tested with the output of the Transkribus Ukraining... model which was already very accurate (0.058). It considerably decreased transcript quality (0.172). None of the models identified the Valamo manuscript base in their sources.  

The successful GPT plan had been: 

\begin{enumerate}
\itemsep0em
\item Transcribe the OCR line-by-line into a plain text file.
\item Identify ambiguous glyphs and compile a list of uncertain readings.
\item Search scholarly sources on Old Church Slavonic paleography for glyph variants.
\item Compare candidate readings with digitized manuscripts and critical editions.
\item Produce a corrected, annotated transcription with justification for changes.
\end{enumerate}

\subsubsection{Post correction and the thinking mode}
The thinking mode in huge LLMs is something which under the hood uses agentic AI-patterns to achieve the task given in a prompt. Instead of inputting the prompt to the LLM directly, often a planner agent is decomposing the task in the prompt into steps which are then executed one by one dynamically. The way in which the thinking mode is
implemented is very different depending on the LLM used. In all, the thinking mode will be computationally more costly and also may be subject to implementation and other adaptation changes (rate limits, etc.) which the user may not be notified of. Resproducibility of results in these modes is therefore also a concern, but this applies to huge LLMs in general, although legacy models are not updated anymore and some download of some versions is also available. 
Using neural network based technology without setting the temperature parameter to 0, the reproducibility issue may apply to VLMs similarly. With the advent of LLMs and their broad usage in CS, the concept of strict reproducibility may have to be relaxed, depending on whether the task permit it. In the case of OCR, as long as the overall CER remains roughly stable, the exact error made may be less important. 

\begin{figure}[H]
    \centering
    \includegraphics[scale=0.5]{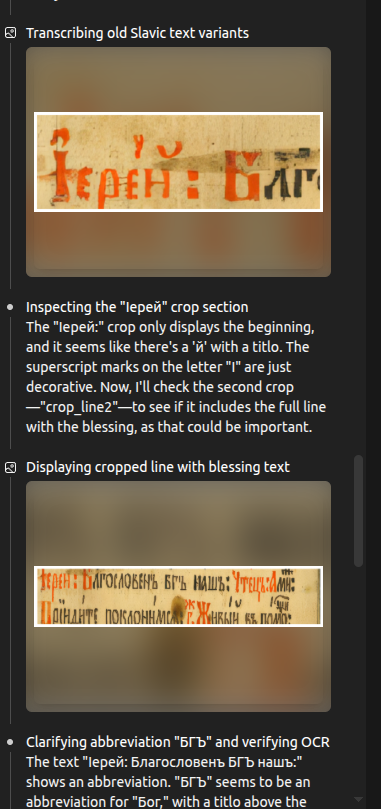}
    \caption{Thinking moment, a look into the inner workings of the thinking LLM.}
    \label{fig:thinking_llm}
\end{figure}

\noindent The thinking mode even though not leading to a better result in our scenario has a feature of displaying the single "thoughts" (the term `thinking' mode bears resemblance to the chain-of-thought term) while users are waiting for a result. The LLM here operates on the image bringing details to the fore. This  may give an additional insight into the models strengths and weaknesses.

In summary, the agentic modes of the models (thinking, deep research, web search) can improve results, but can also deteriorate them partly vastly. 
They present general agentic pipelines not optimized for the OCR-usecase. Designing ones own agentic pipelines with guardrails could be the more feasible approach. The web search can be used for a range of different purposes such as same/similar image search, named entity database search (wikipedia), linguistic ressource search and so forth and then include for instance special websites in the prompt. Thinking modes may help reduce hallucinations. Future research may look into these possibilities. The implentation of these modes is very different from web-model to web-model and might be subject to ongoing changes. Here, instead, building ones own agentic pipelines instead of using precomposed general ones, is put into focus.

\subsection{LLM Post-correction - results}
The post correction prompt was fed to Gemini 3 (GPT respectively) non-thinking or any other mode/tag and the image uploaded alongside. 
\begin{tcolorbox}[
    colback=gray!5, 
    colframe=gray!50, 
    arc=2pt, 
    boxrule=0.5pt, 
    left=10pt, 
    right=10pt, 
    top=8pt, 
    bottom=8pt
]
    \small \textbf{Prompt:} \\
    Given this image of Old Church Slavonic text, and this quite good but far from perfect OCR:
    
    \begin{center}
        \texttt{<transcript>}
    \end{center}
    
    Correct the OCR without expanding abbreviations or adding or deleting anything. Keep orthographic peculiarities also. Consider web sources if necessary.
\end{tcolorbox}
\noindent The phrase to consider web sources was input in order to allow the model, to search for similar data on the web during the OCR-post-correction instead of getting stuck. It was never executed though.

\begin{table}[H]
\centering
\begin{tabular}{l c >{\columncolor{gray!25}}c c >{\columncolor{gray!25}}c c >{\columncolor{gray!25}}c c >{\columncolor{gray!25}}c}
\toprule
Page & Gem & gem1gem & Gem 2 & \textcolor{red}{gem2gem} & Qwen3 & qwengem & T ukr & Tukr-gem \\
\midrule
\rowcolor{gray!15}
1 & 0.079 & 0.042 & 0.055 & \textcolor{green!50!black}{0.040} & 0.088 & 0.038 & 0.071 & \textbf{0.031} \\
\rowcolor{gray!15}
2 & 0.053 & 0.040 & 0.036 & \textcolor{green!50!black}{0.018} & 0.072 & 0.072 & 0.058 & \textbf{0.011} \\
\rowcolor{gray!15}
3 & 0.079 & 0.038 & \textbf{0.020} & \textcolor{green!50!black}{\textbf{0.020}} & 0.070 & \textbf{0.020} & 0.116 & 0.063 \\
4 & 0.117 & 0.093 & 0.154 & \textcolor{red}{0.171} & 0.134 & \textcolor{green!50!black}{\textbf{0.020}} & 0.347 & 0.100 \\
5 & 0.088 & 0.054 & \textbf{0.052} & \textcolor{red}{0.057} & 0.152 & \textcolor{green!50!black}{0.059} & 0.326 & 0.054 \\
6 & 0.154 & 0.114 & \textbf{0.060} & \textcolor{red}{0.104} & 0.191 & \textcolor{green!50!black}{0.066} & 0.504 & 0.095 \\
\rowcolor{gray!15}
7 & 0.104 & 0.089 & 0.053 & 0.049 & 0.177 & 0.070 & 0.193 & \textcolor{green!50!black}{\textbf{0.047}} \\
\rowcolor{gray!15}
8 & 0.132 & 0.040 & 0.040 & \textcolor{red}{0.062} & 0.110 & 0.037 & 0.036 & \textcolor{green!50!black}{\textbf{0.020}} \\
\rowcolor{gray!15}
9 & 0.101 & 0.054 & 0.083 & 0.078 & 0.149 & 0.147 & 0.059 & \textcolor{green!50!black}{\textbf{0.024}} \\
\midrule
Sum & 0.101 & 0.061 & 0.061 & \textcolor{red}{0.065} & 0.127 & 0.063 & 0.172 & \textbf{0.047} \\
\bottomrule
\end{tabular}
\caption{Results of post-corrections via LLMs.}
\end{table}
Overall, the same result as \cite{greif:2025} is obtained, a Transkribus model plus Gemini post-correction worked best overall. However for the different manuscripts different combinations (qwengem and gem2gem) were net best. Yet, upscaling had affected Transkribus models in the second manuscript.

Multi-model input (T gsch+qwen)+gem had been tried, too, but was only slightly better than both alone, more multi-model consensus/voting mechanisms for CS should be researched, for post-correction on LAtin script, consider for instance \cite{nguyen:2021}.

GPT as a post-corrector for T ukr both improved and deteriorated results (omitted here). Notably, in all gem post-corrections only self-correction (gem→gem2) led to deteriorations, which was not the case for the base-prompt results of Gemini corrected by Gemini. This suggests a hypothesis that performance may be bounded by the model’s own accuracy ceiling and correction can cancel out previous gains. Errors are not independent. The confusions in gem2→gem are not primarily due to reintroducing editorial bias, but rather reflect that the same model contributes less new information when applied iteratively. This indicates that fewer steps with a single LLM may be preferable, although excessive cognitive load in one step can likewise be detrimental. Future work should therefore explore adding extra information through prompts as a form of on-the-fly fine-tuning, both directly via few-shot prompting (in-context examples) and indirectly via specialized scientific terminology instructions.

\begin{figure}[H]
\centering
\begin{minipage}{0.8\textwidth}
    \centering
    \includegraphics[width=\linewidth]{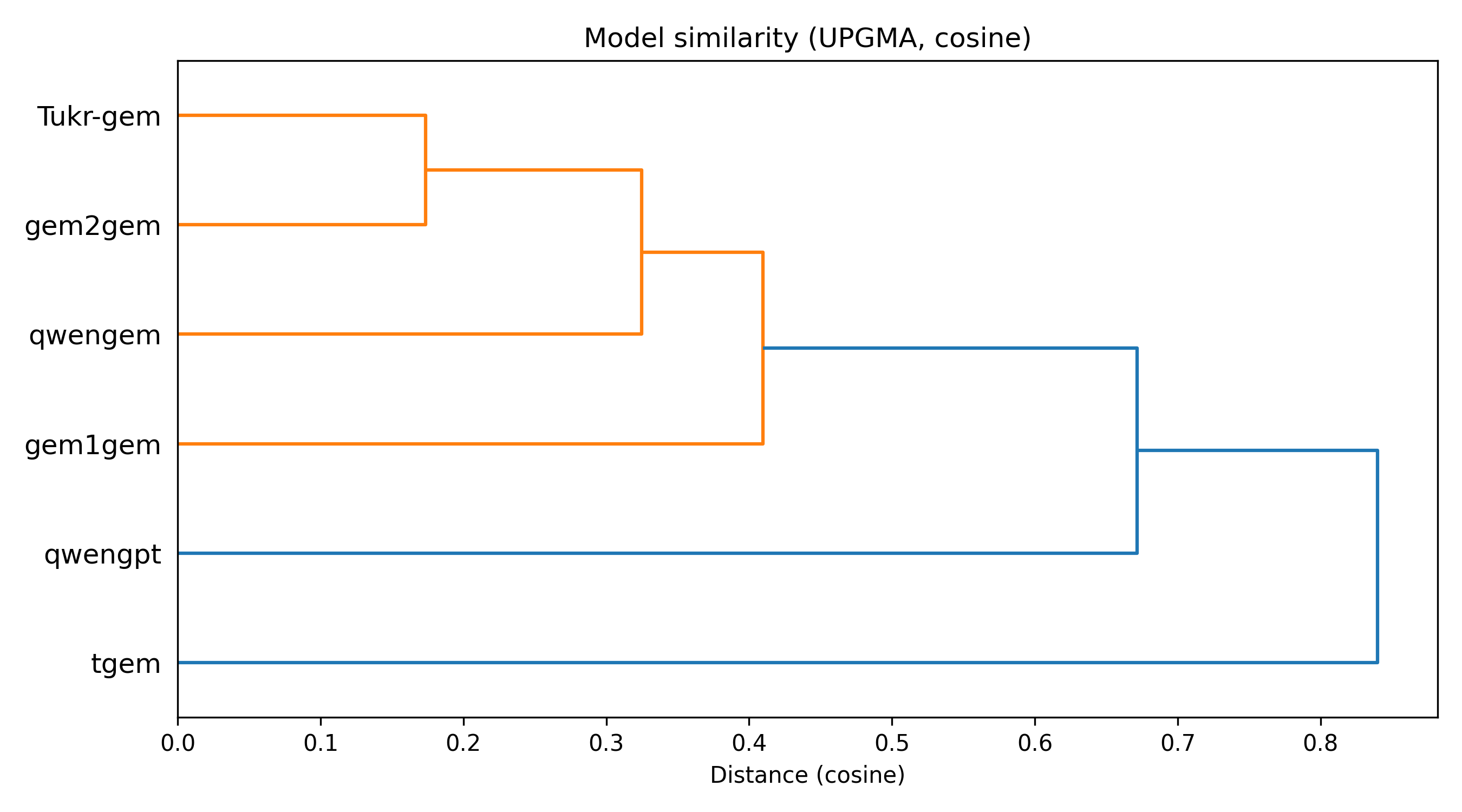}
\phantom{1 Exception: Gem2}
\end{minipage}
\caption{Model similarities of post-corrected. No families discernible.}
\end{figure}

\noindent It can also be seen, that post-correction blurrs the model family borders making the reliability of certain combinations of models harder to predict on the one hand but allowing a better overall result. Combining models which are strong and different may yield the greatest gains.
\cite{kanerva:2025} and \cite{levchenko:2025} on the other hand discuss challenges of deterioration through the application of LLMs.

Ultimately, designing OCR (agentic) AI with LLMs will depend on balancing model diversity, agent roles, and efficient processing pipelines.

\subsection{Three small experiments in agentic AI}

In order to harness the full power of LLMs, one can define so-called agents. In fact, research on LLMs early on has shown, that separating tasks into subtasks increases the LLMs accuracies (chain-of-thought). An agent can be thought of the usage of an LLM for a specialized subtask. However agents or agentic components can also be conventional programs, webservices and so on. Even the escalate-to-a-human ``agent''/or function is part of some call-center applications opening new roads in human-AI collaboration.

For OCR for (Old) Church Slavonic, three small experiments on agentic AI are conducted.

\subsubsection{Agentic OCR-Post-Correction}
Instead of one single post-correction step, three different agents were designed via specialized and sophisticated prompts, in order to allow the model to ``concentrate'' on a special task and become more efficient at it. Some tasks are interdependent and synergies may arise from their joint processing. However some tasks seem to be rather separatable. Here, a first attempt was made, separating post-correction into 3 subtasks:
\begin{itemize}
\itemsep0em
\item a special letter (ligatures, old letters, etc.) diplomatic transcription correction
\item a punctuation and token segmentation correction
\item a diacritic correction
\end{itemize}

As an example,\footnote{All prompts are on the github repositories \url{https://github.com/ArminHoenen/VisualStemma} and \url{https://github.com/ArminHoenen/ocsocr}.} the prompt of the diacritic corrector.\\

\begin{tcolorbox}[
    colback=gray!5, 
    colframe=gray!50, 
    arc=2pt, 
    boxrule=0.5pt, 
    left=10pt, 
    right=10pt, 
    top=8pt, 
    bottom=8pt
]
    \small \textbf{Prompt:} \\
You are an expert in Old Church Slavonic and transcription of christian texts of the 13th to the 18th centuries. You specialize in diacritical marks and non-letter marks apart from punctuation). Visually you are a very able expert for distinguishing image artifacts from true diacritical marks. Analysing the image, correct the OCR with respect to diacritical marks and nothing else. Consider the following typical problems in OCR and diacrits:\\
- correct recognition of superscript letters, (Amen with a superscript 'n') or (Gospodu) with superscript 'c'\\
- breathing marks: if a manuscript uses them, they are often omitted in the OCR\\
- titlos for abbreviations; recognition but also placement may be false\\
- stacking of different marks: Vowel + Breathing + Accent or Base + Superscript Letter + Titlo\\
Output only the corrected OCR. Do not yourself modernize. Do not insert any letters, tokens.
\end{tcolorbox}

The post-correction pipeline was processed through the web surface of Gemini. In order to give access to the previous steps so as to avoid deterioration, everything was processed as one single conversation (where previous and subsequent experiments of course use a new conversation/API call for each extraction). The base image was the non-upsacaled image 1, because in this way the magnitude of the gains of the single steps could be larger. The result of each single step was then fed to the next agent with the instructions and a reupload of the image. 

The result was that the base CER of $0.13$ was reduced to $0.113$ by the letter-corrector, then to $0.091$ by the punctuation and word separation agent and to a final $0.068$ by the diacritic agent. Looking only at the diacritic marks, the first correction agent already reduced the blatant error rate of $0.884$ on diacrits to a mere $0.221$, with the second agent nothing changed and the diacritc agent gained again some accuracy, reducing CER to a final $0.179$ which in absolute terms is still a lot, but on the other hand lower than the ratio of $0.242$, the best model Transkribus Ukrainian was able to achieve on the upscaled image.

This was an encouraging result, especially in this anecdotic example where no deterioration was witnessed at all in any step and where the model never saw the output of any other model.

\subsubsection{Full agentic OCR-Pipeline Sketch}
A full agentic pipeline was generated using pythons LangChain library. This pipeline was generated by GPT5 through a lengthy prompt describing agentic pipeline topology and giving base prompts for single agents and also external tools to be used. The package produced by GPT5 was checked (for instance GPT had slightly altered the single agent prompts) and debugged and then run with two images using the GPT-API.
The approximate topology is seen on the schema:\\

\begin{figure}[H]
\centering
\begin{tikzpicture}[
    node distance=0.4cm and 0.5cm,
    every node/.style={font=\tiny},
    planner/.style={draw, fill=red!10, minimum width=2.5cm, minimum height=0.7cm, font=\scriptsize\bfseries, align=center},
    agent/.style={draw, rounded corners, fill=blue!5, minimum width=2.2cm, align=center},
    tool/.style={draw, fill=gray!10, minimum width=2cm, dashed, align=center},
    output/.style={draw, double, fill=green!10, minimum width=2.5cm, align=center},
    stats/.style={draw, fill=yellow!5, font=\tiny\itshape, align=left, minimum width=3.5cm},
        stats2/.style={draw, double, fill=green!5, font=\scriptsize\itshape, align=left, minimum width=3.5cm},
    arrow/.style={->, >=Stealth, thick}
]

\node[planner] (planner) {Planner Agent\\(GPT-5 Orchestrator)};

\node[above=0.8cm of planner] (input) {\textbf{Input:} OCS Image + Metadata};
\draw[->] (input) -- (planner);

\node[stats, right=1.5cm of input] (stat1) {CER base-prompt GPT \\ non-upscaled image 1: \textbf{0.33}};

\node[agent, left=of planner] (format) {Visual Format Agent\\(JSON 0-9)};
\node[agent, below=0.3cm of format] (ling) {Linguistic Agent\\(Script/Genre)};

\node[tool, right=of planner] (pre) {Visual Tools\\(Upscale, Red2Black...)};
\node[agent, below=0.3cm of pre] (ocr) {OCR Executor\\(Prompt-Guided)};

\node[stats, right=1.2cm of ocr] (stat2) {GPT execution, \\ CER: \textbf{0.225}};

\node[agent, below=1.2cm of planner] (review) {Review Agent\\(Error JSON)};
\node[agent, below=0.5cm of review] (post) {Post-Correction\\(Evidence-based)};

\node[stats2, right=4cm of post] (summary) {Agentic pipeline dynamically\\
uses only needed tools \& models:\\ \\
if borders $\rightarrow$ no Transkribus\\
if red ink present $\rightarrow$ red2black tool\\
...\\
may however deteriorate result};

\node[agent, below left=0.5cm of post] (diacrit) {Diacrit Checker};
\node[agent, above left=0.5cm of post] (consist) {Consistency};
\node[agent, below right=0.5cm of post] (spec) {Special Letter};

\node[output, below=1.2cm of post] (final) {Final Structured OCS Output};

\node[stats, right=1.2cm of final] (stat3) {Post-corrected output (GPT); \\ CER: \textbf{0.19}};

\draw[<->] (planner) -- (format);
\draw[<->] (planner) -- (ling);
\draw[->] (pre) -- (planner);
\draw[->] (planner) -- (ocr);
\draw[->] (ocr) -- (review);
\draw[->] (review) -- (post);
\draw[<->] (post) -- (diacrit);
\draw[<->] (post) -- (consist);
\draw[<->] (post) -- (spec);
\draw[->] (post) -- (final);

\draw[dotted, gray] (input) -- (stat1);
\draw[dotted, gray] (ocr) -- (stat2);
\draw[dotted, gray] (final) -- (stat3);

\end{tikzpicture}
\caption{Architecture and results of the full agentic OCR approach executed with GPT5.}
\end{figure}
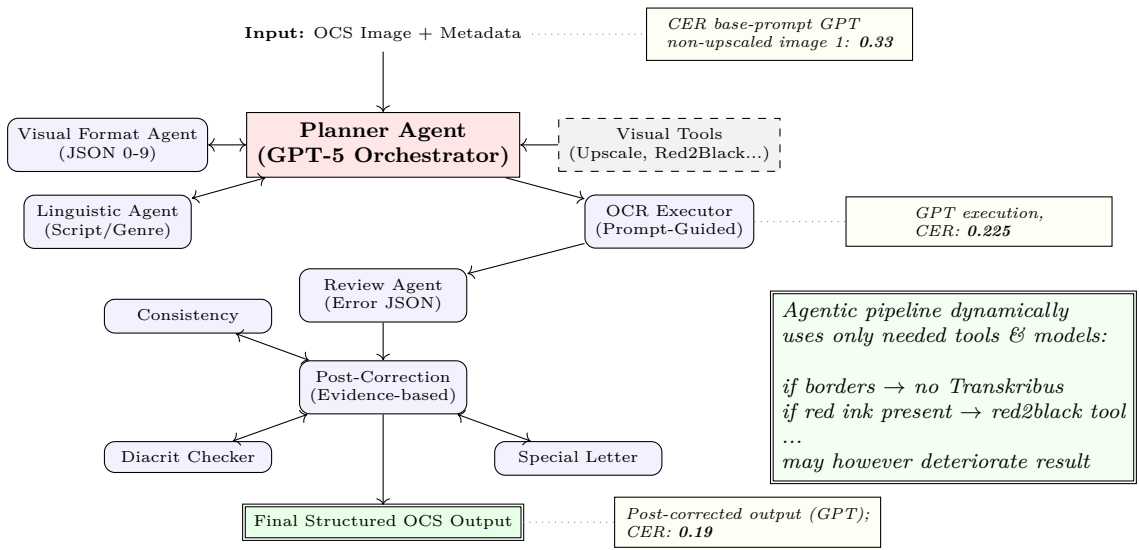

\noindent The OCR if image 7 however deteriorated drastically with the same pipeline. As a sketch, this result shows that such an agentic pipeline can work and be set up relatively quickly. Each agent and tool should then however be optimized thoroughly and guardrails implemented to block deterioration. For instance, comparing the distance of an intermediate output to the result of another external model known to work well for such images and if the distance has considerably increased from the distance of the previous step to that model, beyond some threshold which is to be determined, then the intermediate result is rejected and rolled back to the previous one.

\subsubsection{Agentic RAG for OCR}

Finally, a Retrieval Augmented Generation (RAG) Pipeline was tested. RAG aims to enrich models with external knowledge and has been shown to improve or even enable more accurate results in a wide range of tasks. One usecase of RAG for instance in Machine Translation is to retrieve so-called few-shot examples, that is similar gold standard results together with their inputs on the same task. 
For the OCR task, the RAG process was adapted to operate on lines and to contain triples: image of the line, gold standard line transcript and OCR-engine-OCR of that line.\footnote{For convenience, the line OCR was taken from the OCR of the complete image, not the single line. This can be modelled in this way, when a RAG database is constructed but may involve human linebreak correction.}

In order to test the use case, the data was converted into lines. That is an algorithm cut the upscaled images into lines and this was then human post-corrected, so each line was correct. However, due to overlapping line features such as ascenders/descenders and diacritic marks, the lines were sometimes inferior in information content to their information in the complete image. Then each line was mapped to the corresponding line of the gold standard and the GPT OCR. This was also human-checked. The database was built from images 2-9 and image 1 was tested.\footnote{The presence of images from the same manuscript is something which may not happen in true applications, although it is not excluded. Anyway, the retrieved examples did mostly not come from the same manuscripts.}

The process then operated in the following way: the line image was fed to the LLM (here GPT5 because of the API-access) with a basic extraction prompt. Then the result of this basic extraction was input to the RAG database which retrieved k=3 most similar examples based on Levenshtein distance to the OCR (not the gold standard) of the triples. k is the number of few-shot examples and has been shown in the literature to have a crucial effekt. Which k is best often depends on the task, 1 often is too few, more than 10 is often not adding accuracy. More important however is the quality of the few-shot examples. Here quality is understood in terms of similarity to the current line reading task. The retrieved examples were then, together with the line image, fed again to the LLM, which extracted with this information.

\begin{table}[H]
\centering
\begin{tabular}{lccc}
\toprule
Line & Base CER & Final CER & Base→Context \\
\midrule
1 & 0.200 & 0.133 & 0.000 \\
2 & 0.200 & 0.125 & 0.095 \\
3 & 0.179 & 0.154 & 0.472 \\
4 & 1.000 & 0.875 & 1.333 \\
5 & 0.714 & 0.095 & 1.154 \\
6 & 0.147 & 0.029 & 0.029 \\
7 & 0.091 & 0.000 & 0.045 \\
8 & 0.250 & 0.167 & 0.206 \\
9 & 0.278 & 0.111 & 0.154 \\
10 & 0.368 & 0.158 & 0.400 \\
11 & 0.171 & 0.057 & 0.086 \\
12 & 0.214 & 0.000 & 0.154 \\
13 & 0.212 & 0.242 & 0.212 \\
14 & 0.143 & 0.114 & 0.114 \\
15 & 0.171 & 0.229 & 0.152 \\
\midrule
Sum & 0.258 & 0.151 & 0.235 \\
\bottomrule
\end{tabular}
\caption{Results of the agentic RAG pipeline on single lines of image 1.}
\end{table}

\noindent RAG always improved results (normalized, diacrits challenging). Furthermore in 2 cases, the result was perfect. Those lines were short. OCR results are better if the cognitive load of the model is lower. In this case that means, the fewer letters on the image, the more precise the LLM outcome. Thus, a strategy could be to further visually separate tokens, but whilst lines can relatively well be separated algorithmically, token separation is much harder (and may be impossible with scriptio continua). The distance of the line-OCR when in the whole image and when only from the line image was large. This is certainly due to the malcut descenders/ascenders and diacrits.
This improved the overall CER for the image from $0.258$ to $0.151$. The engine GPT was one of the worst considering the base OCR comparison and the information loss for each line due to cutting further increased difficulty. Despite that, the accuracy gains were large and consistent. RAG pipelines can be included into agentic OCR pipelines as one element or used stand-alone.

\subsubsection{Summing up: Agentic RAG for OCR}
Over all, the three small experiments in Agentic AI for OCR show that there is large potential in applying and configuring agentic pipelines. Prompt optimization can be seen as some on-the-fly-finetuning adapting large LLMs very much. Adding external information from other models can help improve results as much as task separation (reducing cognitive load), the intergration of external tools such as image enhancement or corrective web-search and within model few-shot examples (RAG). A challenge in the use of large LLMs as OCR-systems is possible deterioration which seems less probably with RAG, but for which mechanisms may be found to limit it. A second challenge is the reproducibility when using online models. Finally, the more complex an AI-OCR system gets, the more time, energy and infrastructure it will consume.
For church slavonic, the chances lie in a dynamic process of auto-adaptive pipelines for very diverse sets of input (image qualities, deterioration of the material, time step, subgenre, ...) increasing overall accuracy and maybe cancelling the need of compiling specialized models and corpora for subclasses of manuscripts.
Simple OCR-post correction as investigated in the literature a little is by far not the only and maybe not the most successful application scenario of large LLMs to OCR.

\subsection{Unique Selling Propositions? Image-first and multiple-script input}

\begin{figure}[H]
    \centering
    \includegraphics[scale=0.05]{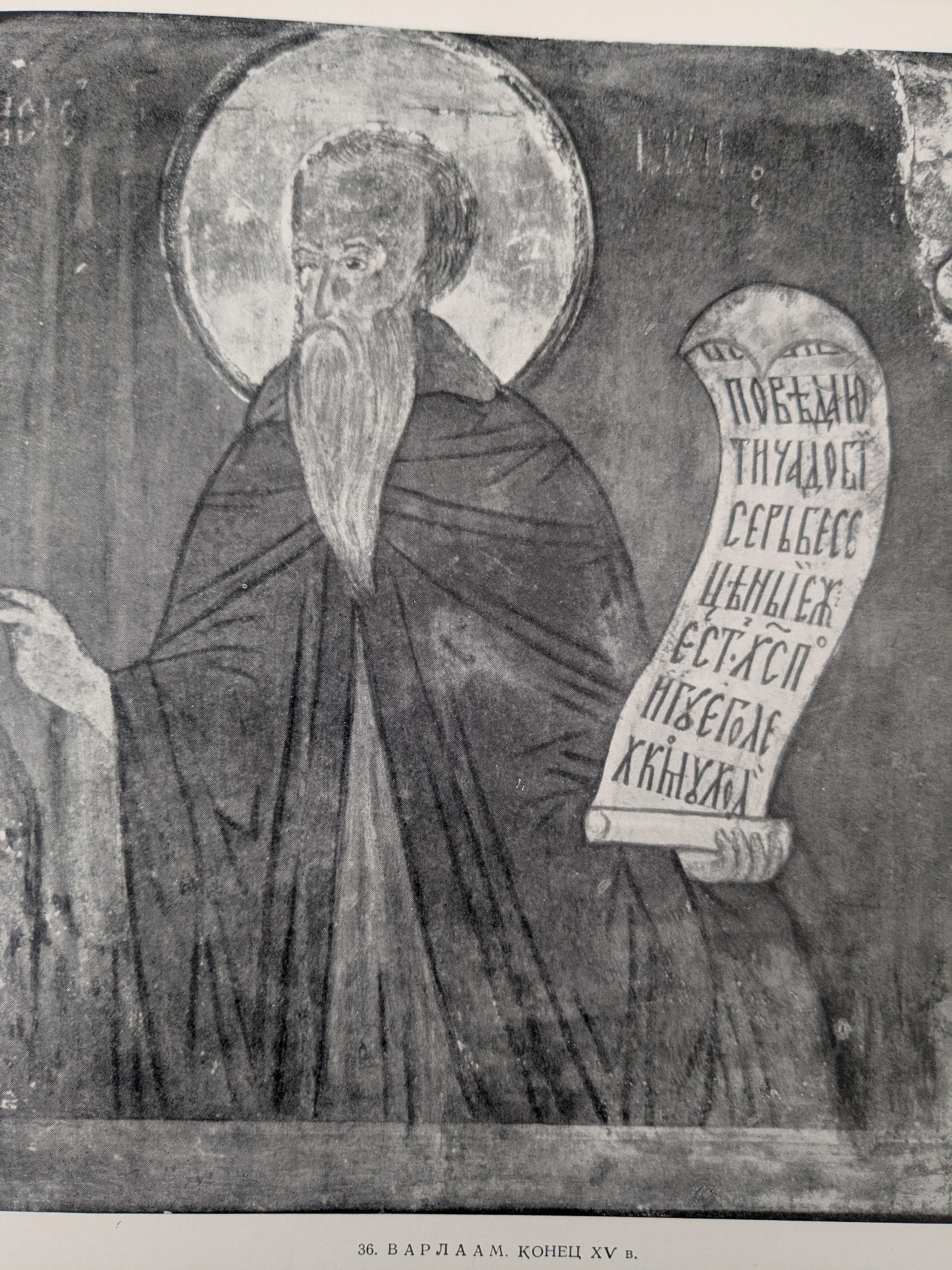}
    \caption{Icon photo of photo, book O. Zonowa, \textit{The treasures of the Kremlin}. Moscow:1963, No. 36. Warlaam, end of the 15th century. Not enhanced with upscayl.}
    \label{fig:icona}
\end{figure}

\begin{table}[H]
    \centering
    \caption{OCR Model Comparison with Error Highlighting}
    \label{tab:ocr_highlight}
\includegraphics[scale=0.5]{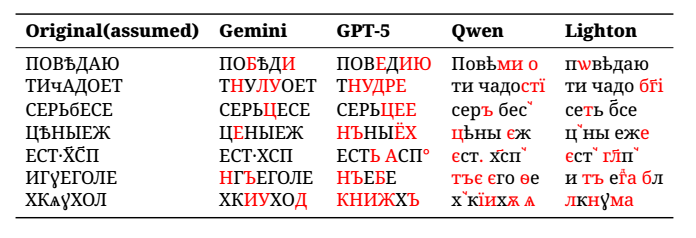}
\end{table}
Basic extraction prompt.
Ranking: Gemini (half the errors of lighton), lighton, qwen, gpt.

One can see that the Gemini Model was most successful but still had room for improvement. Another interesting aspect is that the LLM models seem to prefer Upper Casing here, whilst in the texts above, they are not different from the other models in preferring lower case in general.

Finally, I tested a coin with two scripts. The script was totally bent. If scripts are mixed momentarily LLMs seem to be the only feasible solution as the other trained or finetuned models are script dependent, if not VLM based.
 
The languages being modern Russian and English, Gemini and GPT5 extracted 100\% correct: spass na krowi- SAVIOR ON THE SPILLED BLOOD
(and the flipside ST.-PETERSBURG ALEXANDER II).
Lighton only transcribed the cyrillic and failed at that (сїа си̏ кротиⷭ҇), but bent text seems to work even though the lighting of the coin was likewise particular.
Qwen instead interpreted the Latin text as Church Slavonic. However, bent text is no problem to VLM-based OCR-machines. Tesseract was not able to recognize text on the image, but also for Transkribus it was too far from the training.
The abilities of large LLMs to extract text in these difficult conditions and without extra finetuning or prompt optimization whatsoever render them allrounders, suitable for very diverse digital collections (or smaller ones which rather than containing many objects of one type contain a range of different objects [smaller monasteries, museums etc.]).
LLMs possess some unique selling propositions also because they are multilingual. For low ressource scenarios, where experts and personell are rare, they might produce useful working output.

\section{Forgeries}
The LLM models do not only read and extract text from images, they are able to produce images. What can that tell us about their OCR capabilities if anything? 
The models tend to be able to reproduce shapes from their training (statistical average) and not to be able to produce what is very different from their trainng data range unless uploaded as additional input. So forging without additional data input may tell us roughly what they might have seen most and what maybe not or not at all.

A central question arises from the observed limitations in OCR and post-correction: if models struggle with these tasks, are they similarly limited in the production of forgeries?

At least you can get an idea by trying to use the same models you intend to use as OCR engines beforehand in tests of image generation so as to get an idea what may work well and what not, or in order to compare models.

\subsection{Production Characteristics and Constraints}
Current generative models exhibit specific patterns when attempting to replicate (Old) Church Slavonic or Cyrillic manuscripts:
\begin{itemize}
    \item \textbf{Spacing Dependency:} Models rely heavily on spaces; they often fail to omit them even when explicitly instructed to avoid spaces between words and produce scriptio continua. That might be due to the fact, that modern digitization tends to always add them.
    \item \textbf{Typographic Uniformity:} There is a tendency to produce "print-like" identical letters rather than handwritten script variation. Here, Gemini was able to move beyong the limitation thanks to an uploaded image.
    \item \textbf{Hallucinated Diacritics:} Diacritical marks are often generated inconsistently or hallucinated entirely.
    \item \textbf{Structural Errors:} Models may produce messy or incorrect letter forms  where a human would likely correct, cancel out or simply not err (though the ways of human error are ...).
\end{itemize}

\noindent Despite these hurdles, preliminary tests with Gemini-base (fewer than 10 attempts) suggest that convincing forgeries are possible even without the user possessing knowledge of (Old) Church Slavonic. The "first-best" prompt rarely succeeds, but iterative refinement and strategic image degradation—simulating ("old photograph") possibly combined with human post-editing can produce high-quality and maybe even relatively convincing specimen.

\begin{figure}[h]
    \centering
    \begin{minipage}{0.45\textwidth}
        \includegraphics[width=\textwidth]{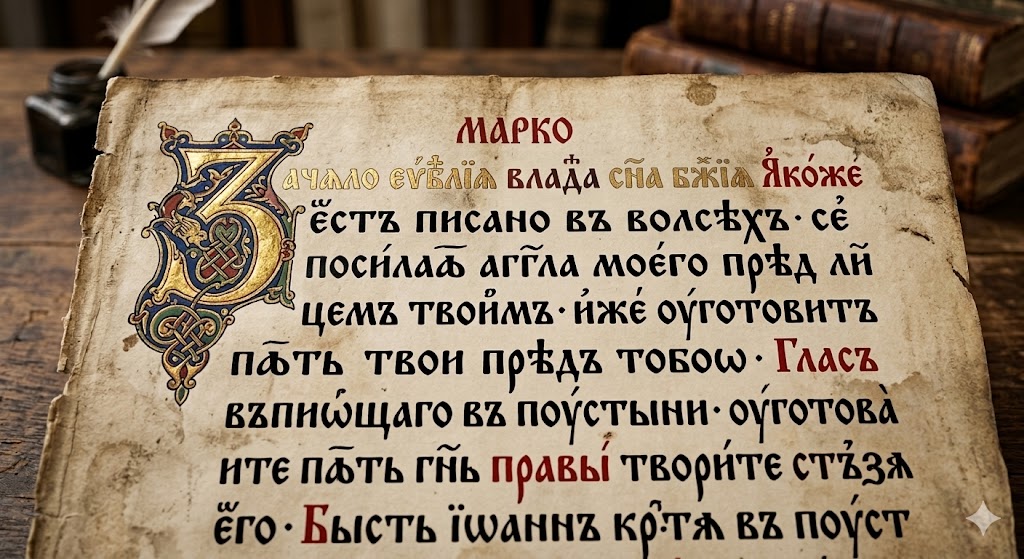}
    \end{minipage}
    \hfill
    \begin{minipage}{0.45\textwidth}
        \includegraphics[width=\textwidth]{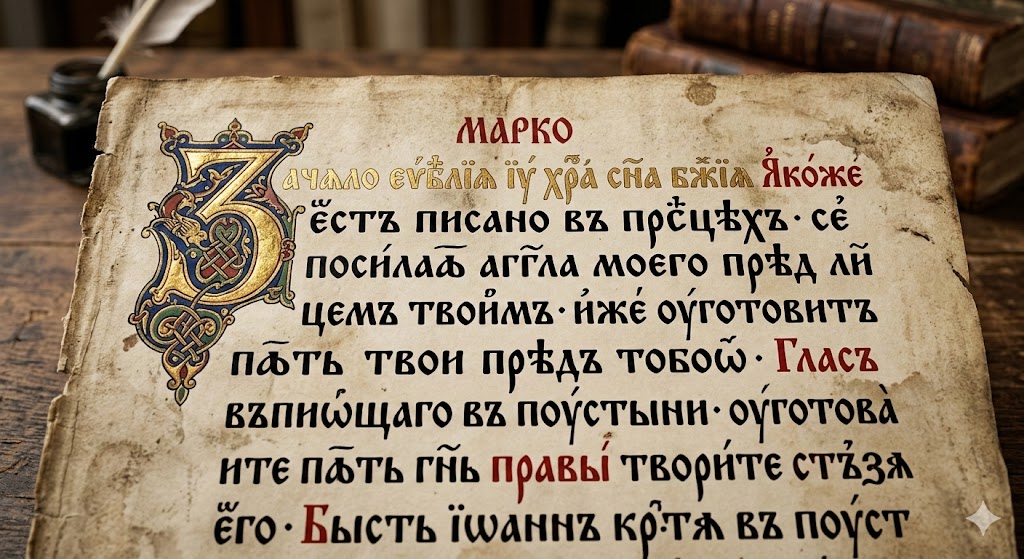}
    \end{minipage}
    \caption{Left: Forged by Gemini by substituting Jesus with Vlad and Prophets with Sorcerers from the previous Gemini forgery on the right.}
\end{figure}

\subsection{Analyzing the "Model Bubble"}
When provided with an example image, character variation improves, as the letters are no longer "print-exact" twins. This suggests that in-context learning partially relieves the model's internal constraints. 

\begin{figure}[h]
    \centering
    \includegraphics[width=0.6\textwidth]{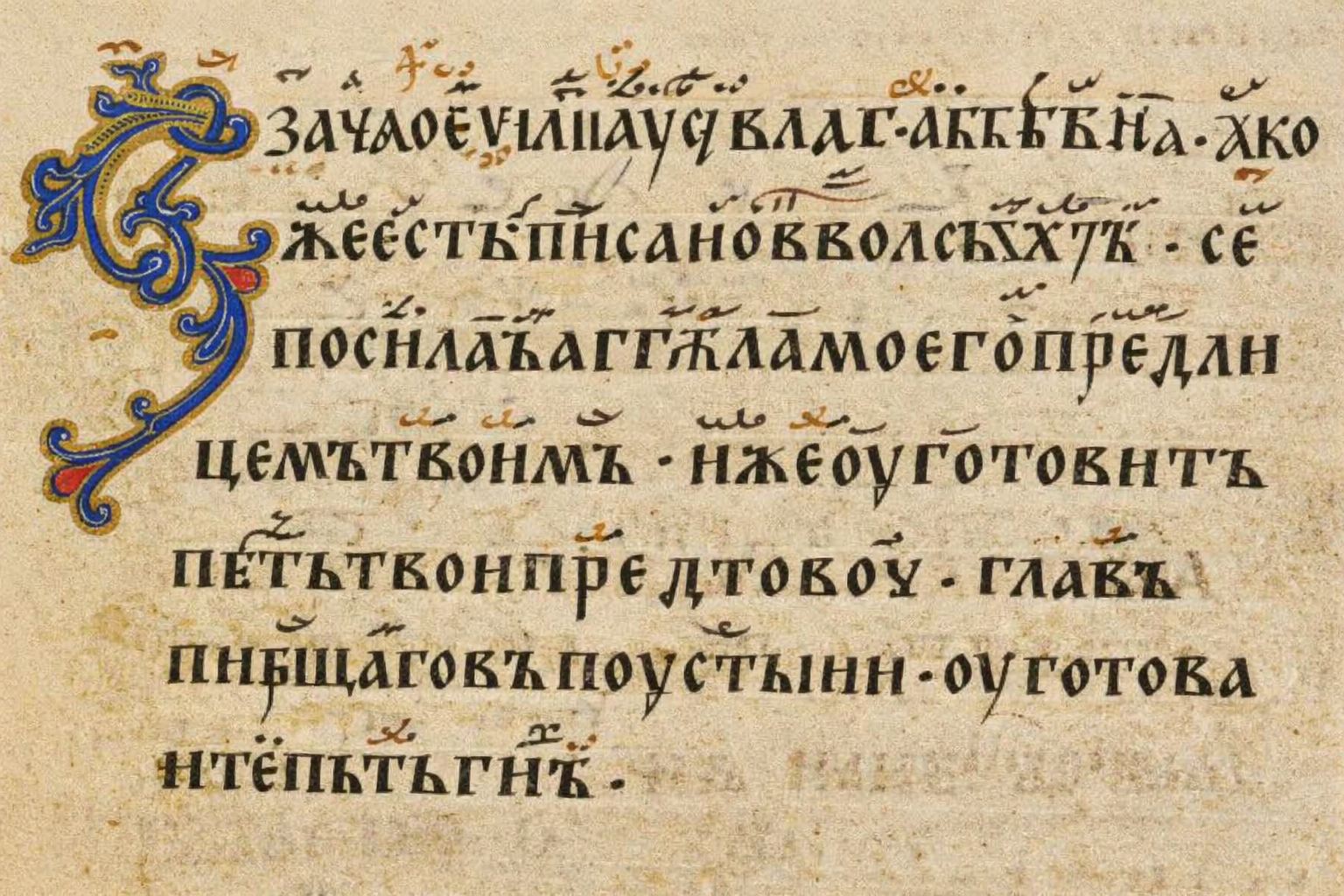}
    \caption{GPT generated with an example image; note that letters are not print-exact twins.}
\end{figure}

\noindent Interestingly, models may be more adept at OCRing their own generated content—staying within their "bubble"—though they might identify a forgery once the source of the constraint is understood. When performing OCR on the forged \texttt{gemini2.jpg}, the model produced a transcription which possibly had fewer errors than the model produces OCR errors on real-world data (almost none), yet notably, the diacritics were not OCRed and uppercase letters were produced for title and first part of the manuscript (remember the icon). Finally, the modernized letterforms are being used. Maybe one could use the CER of a forgery with an uploaded image to measure how much the image actually is able to push the model towards certain features.

\subsection{Obfuscation and Unrestricted Models}
The detection of forgeries (deep-fake recognition, debunking) might rely on dealing with the aforementioned challenges: spacing, diacritics, and handwriting variation. However, such features and recognizability can be masked by obfuscating image quality.

\begin{figure}[H]
\begin{center}
    \begin{minipage}{0.45\textwidth}
        \includegraphics[width=\textwidth]{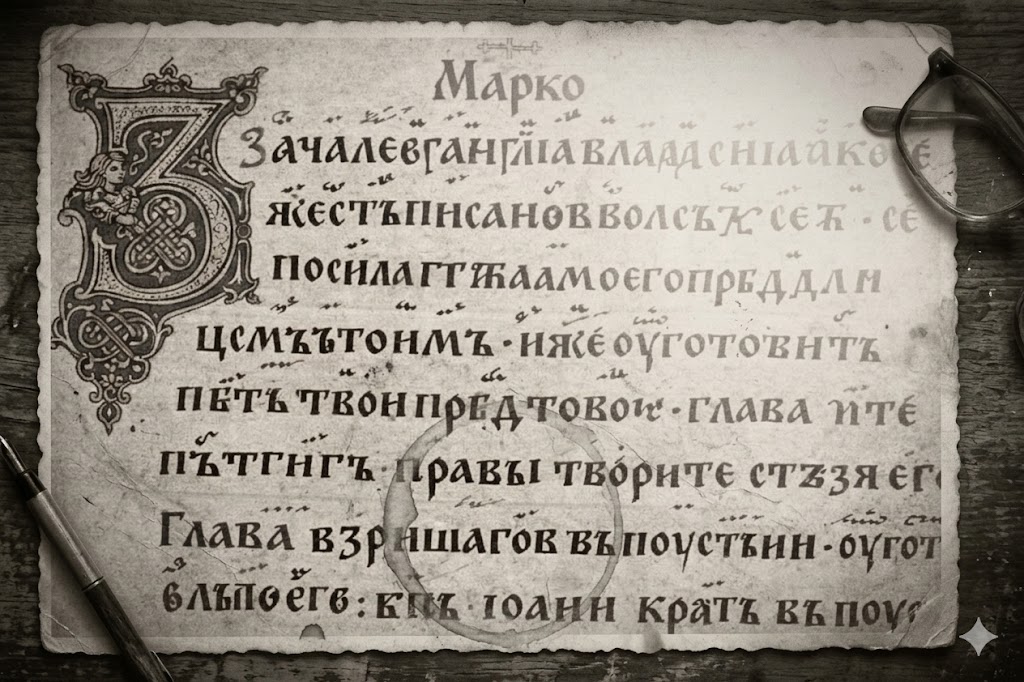}
    \end{minipage}
    \hfill
    \caption{Obfuscating forgeries by making the image less sharp, degraded, badly lighted and the letters smaller.}
    \end{center}
\end{figure}

\noindent The path toward sophisticated image-text forgeries is further expanded by unrestricted models, which bypass certain constraints (like producing nudity) and allow for more diverse pathways in content generation.

\begin{figure}[H]

    \begin{minipage}{0.3\textwidth}
    \begin{center}
        \includegraphics[width=\textwidth]{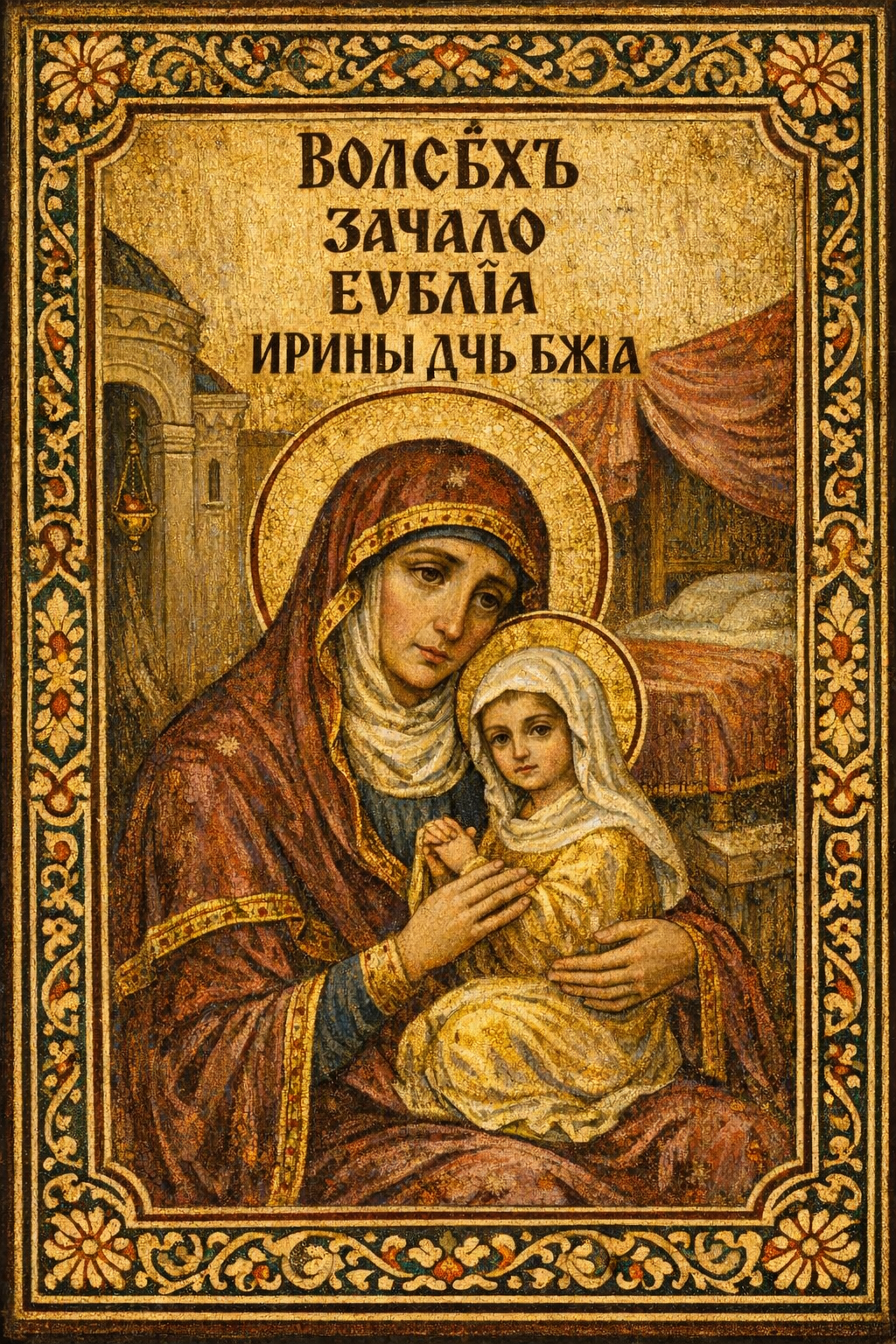}
    \end{center}
    \end{minipage}
    \begin{minipage}{0.3\textwidth}
    \begin{center}
        \includegraphics[width=\textwidth]{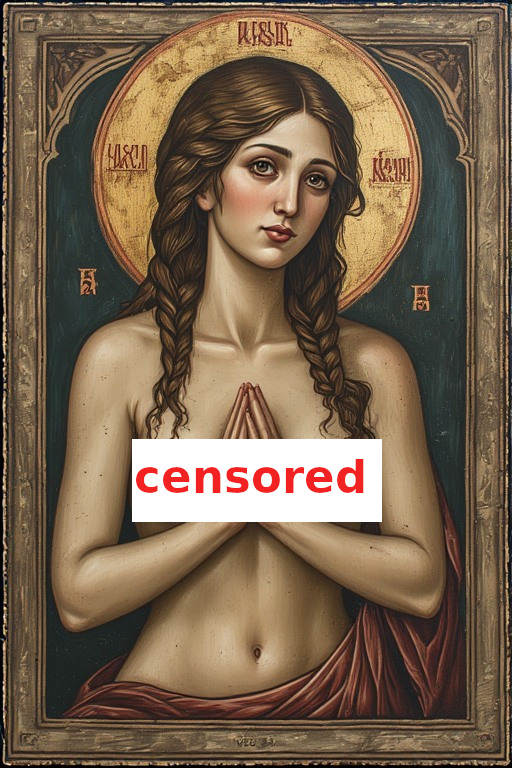}
        \end{center}
    \end{minipage}
    \caption{Examples of generated forgeries, including output from a post-edited unrestricted model.}

\end{figure}

\noindent Finally, forged images may support theological argument of any kind or other more prophane visions. Old church slavonic writing seems to be present enough in the training of large models. Here, we use it for scientific purposes. Having a look at the original is nowadays probably always a good idea, fotos can be printed of course.

For understanding OCR better, forgeries can be made, they should not have to be offensive or blasphemous, yet for understanding the limits of what is possible, some borderline content might be impossible to avoid.

\section{Image-native stemmatology}
A stemma codicum is a visualization of the copy history of witnesses of a textual tradition. Although admixtures of texts (contamination) and other phenomena imply that a Directed Acyclic Graph is not always the most appropriate way of visual representation, by and large stemma generation has become a standard, especially in the digital age, where the generation is relatively straightforward and bio-informatic tools facilitate stemma generation. The generation needs digitizations of the witnesses, for that matter all witnesses or at least a critical apparatus, allowing for the automatic reconstruction of those.
Visual models long struggled with text and OCR basically is visual processing of letters, so it does not surprise that so far, no method has been proposed, which generates a stemma from image information alone. The first idea which may come to mind would be to vectorize the images of the digitized manuscripts and then compute pairwise cosine-distances of the images or so. However, letter size, font and format, lighting, ink color, illustrations, illuminations, degradation and many other factors would influence these distances. It is apriori implausible that such a method should work.
Here, another approach is pursued. Before, however, we investigate, whether the OCR alone is not enough for stemmatology. If OCR were perfect, then no human transcription or post-editing would be necessary. 

\subsection{OCR and textual distances - the Gospel of Mark}
For three witnesses in handwritten Old Church Slavonic from the 14th century, we conduct OCR and then measure pairwise distances, comparing those distances with the gold standard distances. 
Those witnesses are the first pages of:
\begin{itemize}
\itemsep0em
\item London, British Library, Add MS 39627
\item Munich, Bayerische Staatsbibliothek, Cod. slav. 1
\item Munich, Bayerische Staatsbibliothek, Cod. slav. 25
\end{itemize}
The text has been made sure to end at the same point in text by otherwise cutting additional content. 
Using Gem2, OCR worked considerably worse than on our previous examples which in part could be due to the much older forms and scriptio continua which erases some of the benfits of contextual processing. The OCR has spaces.
\\
\begin{table}[H]
\centering
\begin{tabular}{lcc}
\toprule
MS & Lev Dist & CER \\
\midrule
tsar & 35 & 0.223 \\
slav1 & 34 & 0.221 \\
slav25 & 30 & 0.195 \\
\bottomrule
\end{tabular}\\
\caption{The distances of the OCR of the three manuscripts.}
\end{table}
\noindent The distances from the gold standard are slightly different for each manuscript. This is a random effect, but as long as distances are larger, random fluctuations can be more pronounced.
Now, computing pairwise distances on the basis of the OCRs leads to the following results [gold standard distances in brackets]:\\
\begin{table}[H]
\centering
\begin{tabular}{lccc}
\toprule
 & tsar & slav1 & slav25 \\
\midrule
tsar & -- & 28 [3] (21 [1]) & 26 [1] (36 [3]) \\
slav1 & -- & -- & 26 [2] (27 [2]) \\
slav25 & -- & -- & -- \\
\bottomrule
\end{tabular}\\
\caption{Pairwise distances, with gold distances in parenthesis and ranks in brackets.}
\end{table}
Apart from OCR outputs being net-closer to each other, no matter what OCR method or model, 
the true relations can be masked. Here, the truly lowest distance becomes the largest. Such a distortive effect may be called OCR-relation-distortion. This is unfortunate for stemmatology, because it means ALL versions, even minor ones, must be fully human-post-corrected before stemmatology is feasible. The scientific community is aware and has therefore never attempted to compute final stemmata based on raw OCR-outcomes. 
However, algorithms have been assessed towards their robustness and bias, under more using OCR-distortion. A well-known robustness increasing method in bio-informatics is bootstrapping. \cite{dumitrescu:2025} argue for uncertainty values to be used but also for 'abstention' where uncertainty is too large. Algorithms self-certainty about their results unfortunately does not directly correspond to the correctness of their results. 

Even with LLMs, we have seen that the error rate will not become 0. Especially scriptio continua and diacritical marks make transcription more difficult for Old Church Slavonic.

The toll on low resource language or domain scenarios is naturally larger. Here, a purely visual method could relief the need for rare experts, whose time is probably better spent other than in post-correcting minor versions. In the worst case, experts are not available at the required time and place of funding. A purely visual method would allow a cross-linguistic and annotationless stemma-computation. But, how should this be achieved?

\subsection{The Purely Visual Stemmatic Method}
The idea is to abstract away from the image by a multi-step-pipeline. Starting from page digitizations showing exactly the same section of a text (ideally the digitization of an entire manuscript).

\begin{itemize}
\itemsep0em
\item automatically separate out all glyphs
\item cluster the glyphs: the goal is to obtain clusters corresponding to letters, which is why a cluster algorithm such as k-means with a k corresponding to the number of letters may be a good initial hypothesis; k then would be some additional parameter, a human should predetermine and input
\item map clusters for each manuscript pair and then compute a distance based on the difference of the distributions of glyphs among clusters which with perfect glyph separation and clustering would correspond to letter distributions
\item from a distance matrix, there are various methods to compute a stemma
\end{itemize}

Besides many caveats and details (such as how to deal with lacunae, majuscules vs. minuscules etc.), the distances obtained from the true letter distributions should correspond to stemmatically relevant distances.

\subsection{Literature}
Machine learning approaches for corpus-wide analysis of historical sources have recently been explored using unsupervised atomization–recomposition strategies, where documents are decomposed into recurring visual primitives and recombined into higher-level similarity representations without full transcription. \cite{eberle:2024} apply this method to 359 early modern printed astronomy works (1472–1650; about 76,000 pages), using vision models to detect graphical “atoms” (e.g., digits, symbols) and construct bag-of-features–like descriptors that capture content-specific similarity rather than sequential OCR output; this demonstrates that large-scale grouping of historical sources can be achieved without heavy labeling and supports structured image-feature approaches to visual stemmatics.

\subsection{Letter Distribution Distances for Perfect Cases}
In order to simulate a perfect case, the first script size 12 page of each witness of the artificial tradition of Parzival (\cite{Spencer:Davidson:Barbrook:Howe:2004}) has been transformed into a letter distribution map. This has then been used to compute pairwise manuscript distances. These distances have then been compared to the gold standard Levenshtein distance matrix of the gold standard. The spearman rank correlation was $0.9$ which strongly suggests that the information obtained from the letter distributions is largely equal to a textual distance such as Levenshtein. This is good enough to apply the method, even though it is clear that places of variation (and so-called Leitfehler) are never defined in terms of letters but tokens or wordings and a letter based method can probably never compensate the missing information. Visually cutting tokens instead of glyphs would be the logical alternative, but then, scriptio continua would not be processable. 
This is another of the many details, which will have to be rigorously investigated by future research. Meanwhile, the information from the letters appears good enough to generate non-random distance outputs and thus a fair approximation especially in low resource scenarios, where maybe no other method or expert is available.

\subsection{A first application on the Gospel of Mark }
In order to extract the glyphs, visual processing was used, scripts were generated using vibe coding with Gemini and GPT and iterative testing. 

\subsubsection{Visual Glyph Separation}
The first algorithm performs classical connected-component-based character segmentation: the image is converted to grayscale and binarized using Otsu’s global thresholding to separate foreground strokes from background \cite{otsu:1979}; morphological opening with a small structuring element removes noise and bridges nearby pixels following standard morphological filtering practice \cite{serra:1983}; external contours are then extracted using OpenCV’s contour-based connected-component detection and filtered by bounding-box size to suppress artifacts \cite{opencv_contours}; finally, bounding boxes are sorted by approximate line (Y-binning) and horizontal position, and padded crops are saved as individual character segments, forming a typical OCR preprocessing pipeline implemented with OpenCV and NumPy \cite{opencv_library,numpy}.

\subsubsection{Glyph Clustering}
The script performs unsupervised image grouping by extracting deep visual embeddings with a pretrained ResNet-18 convolutional neural network (final classification layer removed) to obtain fixed-length feature vectors from normalized 224×224 RGB inputs \cite{he:2016}; these vectors are computed in inference mode using PyTorch and standard ImageNet preprocessing, producing semantic descriptors commonly used for transfer learning \cite{sharif:2014}; the resulting feature matrix is then clustered with k-means, which partitions samples by minimizing within-cluster variance in Euclidean space \cite{lloyd:1982}; k is a parameter uniformly passed to all manuscripts. Finally, images are assigned to clusters and copied into separate directories, yielding unsupervised grouping of visually similar symbols based on deep feature similarity.

\subsubsection{Cluster Mapping, Discarding of Disjunct and Distance Computation}
The third algorithm in the pipeline compares clustered symbol sets averaging them to obtain cluster centroids in feature space; pairwise cosine similarity between centroids forms a similarity matrix measuring angular proximity of clusters; a one-to-one correspondence between clusters is then obtained by solving the linear assignment (Hungarian) problem on the similarity matrix (converted to a cost), yielding the globally optimal matching \cite{kuhn:1955}; finally, low-similarity matches are discarded and the remaining pairs are weighted by relative cluster frequencies to compute an average distance between manuscripts, producing a deep-feature-based structural similarity measure. The percentage of discarded mapped clusters is a parameter. It has been set to 10\%, which was empirically tested.
\[
d(M_1,M_2)=
\frac{1}{n}
\sum_{\text{matched clusters}}
\left| f_1 - f_2 \right|
\]
where n is the number of clusters and f the frequencies.

\subsubsection{A suspended improvement - alphabet aware mapping}
An alphabet-guided experiment was undertaken, where clusters where not matched directly between the manuscripts, but pivoted through a canonical alphabet reference. Typical letters such as from alphabets provided on the web (here the wikipedia article on the church slavonic alphabet) were used.
Not the 10\% worst clusters of the original manuscript mappings were discarded but those worsely aligning with the alphabet.
This approach follows prototype-based clustering and assignment filtering strategies commonly used for noise reduction in unsupervised visual grouping. In principle, using the alphabet as a pivot across manuscripts should suppress non-letter clusters before pairwise comparison, but preliminary results showed instability due to embedding noise and visual ambiguity (e.g., C matched to K), consistent with known sensitivity of deep-feature nearest-prototype matching under small-sample conditions.

\subsection{Results}

\begin{table}[H]
\centering
\begin{tabular}{l c}
\toprule
Pair & Visual distance \\
\midrule
slav1--tsar & 0.0157 [1] (1) \\
slav1--slav25 & 0.0172 [3] (2)\\
slav25--tsar & 0.0168 [2] (3)\\
\bottomrule
\end{tabular}
\caption{Visual distances obtained from the method.}
\end{table}
The result for the Gospel of Mark corpus in this case was still slightly better than that obtained by the distances from the raw OCR. 

Applying the visual algorithm to the artificial tradition corpus, the spearman rank correlation with the gold standard Levenshtein distances was a mere $0.55$ when discarding 10 \% of the worst cluster mappings, but dropped to around $0.2$ when using all clusters. Inspecting glyph seperation and clustering, it became clear that both are still very imperfect. Instead of single glyphs, sometimes tokens have been obtained, diacritical marks and other parts of letters were obained. This is one of the difficult details, since it implies the question of how to choose k. Of course, the glyph level errors propagated into the clustering step, which had its own imperfections on top. Many of the clusters were impure and some mixed certain similar letters. Especially, the glyph separation algorithm should be optimized or a human separation step could be imagined. That would shift annotation effort from OCR-text-correction to glyph separation or cluster correction (which would still be easier tasks).\\

\begin{figure}[H]
\centering

\includegraphics[scale=0.3]{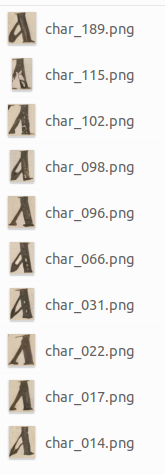}
\includegraphics[scale=0.3]{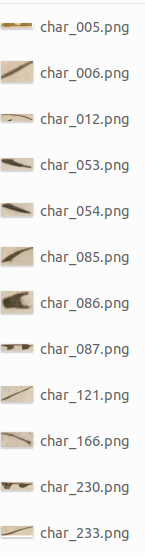}
\caption{Cluster example snapshots, left good cluster, right bad cluster.}
\end{figure}

\subsection{Roman de la rose}
A second experiment was run on 4 manuscripts of the 'Roman de la Rose'. The \emph{Roman de la Rose} is a thirteenth-century allegorical dream-vision poem written by Guillaume de Lorris and continued by Jean de Meun, and it survives in hundreds of illuminated manuscripts across European collections. Here, the following 4 witnesses have been used:

\begin{itemize}
\itemsep0em
\item cam: \textbf{Cambridge, UL MS Gg.4.6 (Roman de la Rose)} — A richly illustrated early-14th-century Parisian copy of the allegorical Old French dream-vision poem by Guillaume de Lorris and Jean de Meun, one of the most widely transmitted literary works of the Middle Ages.\footnote{\url{https://cudl.lib.cam.ac.uk/view/MS-GG-00004-00006/15}} \cite{ } 

\item bod: \textbf{Oxford, Bodleian Library (Digital Bodleian, object fdc348b0…)} — A late-medieval illuminated manuscript of the \emph{Roman de la Rose} preserved in the Bodleian collections, reflecting the poem’s extensive fifteenth-century manuscript circulation.\footnote{\url{https://digital.bodleian.ox.ac.uk/objects/fdc348b0-c5e4-4e97-843f-4395e4feb2cc/}}

\item bod2: \textbf{Oxford, Bodleian Library (Digital Bodleian, object c4cf7c85…)} — Another Bodleian copy of the \emph{Roman de la Rose}, written in Middle French and attributed to the tradition of Guillaume de Lorris and Jean de Meun, representing a later manuscript witness.\footnote{\url{https://digital.bodleian.ox.ac.uk/objects/c4cf7c85-8c5f-4b8f-98ec-53a3509a9e07/surfaces/9319f69d-4247-4ca3-9918-811a7f3a5835/}}

\item illi: \textbf{University of Illinois manuscript (Roman de la Rose)} — A digitized medieval French manuscript witness of the allegorical courtly-love poem, part of the broader corpus of over 300 surviving copies produced between the thirteenth and sixteenth centuries.\footnote{\url{https://digital.library.illinois.edu/items/5d57ccb0-05a0-013d-4d2c-02d0d7bfd6e4-a}}
\end{itemize}

Here again, the first pages were screenshot and additional text was masked, so all witnesses ended at a corresponding place. Then the algorithmic pipeline was exectuted and from the distance matrix, a stemma was generated using neighbour joining \cite{Saitou:Nei:1987}. Groupings of the manuscripts from the literature could correspond to this stemma. For more in depth insights into that tradition, consider \cite{sympson:2012}.

Glyph separation errors and clustering could have affected all manuscripts in the same way leading to feasible results.

\begin{table}[H]
\centering
\begin{tabular}{l c}
\toprule
Pair & Visual distance \\
\midrule
bod\_adjust -- cam\_adjust   & 0.013872 \\
cam\_adjust -- illi\_adjust  & 0.015888 \\
bod2\_adjusted -- cam\_adjust& 0.020303 \\
bod2\_adjusted -- illi\_adjust& 0.021984 \\
bod2\_adjusted -- bod\_adjust& 0.022026 \\
bod\_adjust -- illi\_adjust  & 0.026703 \\
\bottomrule
\end{tabular}
\caption{Distances obtained through method.}
\end{table}

\begin{figure}[H]
\centering
\includegraphics[scale=0.5]{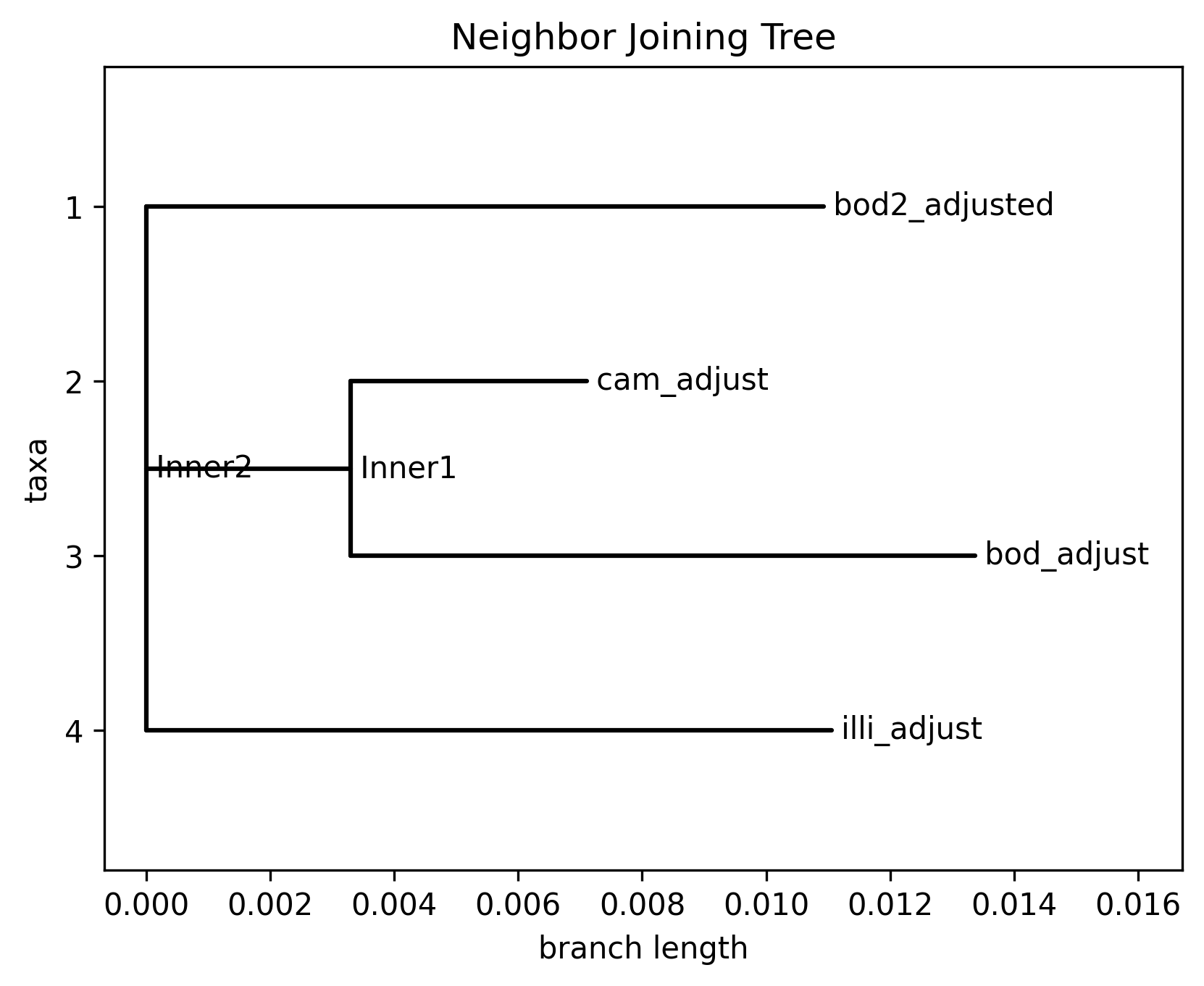}
\caption{An unrooted topology obtained by Neighbour Joining for the 4 manuscripts.}
\end{figure}

\subsection{Conclusion}
Whilst purely visual stemma generation has been shown to be possible, many details must be elaborated and especially automatic glyph separation optimized. The decision as to what would count as a smallest unit of clustering (how to deal with diacritic marks, ligatures etc.), how to deal with lacunae and other visual, non-deterministic artifacts or elements of the manuscripts, how to incorporate meta-data into the stemma generation process, as do \cite{hyytiainen:2026}, and presumably many more are details, which would precede larger experiments with complete manuscripts. The method itself, despite the many imperfections could be the only available stemmatological assessment for low resource corpora with severe lacks of experts or textual data.
Data, scripts and results are released on \url{https://github.com/ArminHoenen/VisualStemma} and \url{https://github.com/ArminHoenen/ocsocr}.

\bibliography{HoenenPromFIN.bib}


\end{document}